\documentclass[10pt,twocolumn,letterpaper]{article}

\usepackage{iccv}
\usepackage{times}
\usepackage{epsfig}
\usepackage{graphicx}
\usepackage{amsmath}
\usepackage{amssymb}
\usepackage[pagebackref=true,breaklinks=true,letterpaper=true,colorlinks,bookmarks=false]{hyperref}

\usepackage{booktabs}
\usepackage{multirow}
\usepackage{leftidx}
\usepackage{bbm}
\usepackage{floatrow}

\usepackage{algorithmic}
\usepackage{textcomp}
\usepackage{siunitx}
\usepackage{tablefootnote}
\usepackage{footnote}
\usepackage{caption}
\usepackage{makecell}
\usepackage{pifont}% http://ctan.org/pkg/pifont

\newcolumntype{P}[1]{>{\centering\arraybackslash}p{#1}}
\newcolumntype{M}[1]{>{\centering\arraybackslash}m{#1}}

\DeclareMathOperator*{\argmax}{arg\,max}

\newcommand{\OurMethod}{MAXI\xspace}

\newcommand{\myparagraph}[1]{\vspace{2pt}\noindent{\bf #1}}

\makeatletter
\def\blfootnote{\gdef\@thefnmark{}\@footnotetext}
\makeatother

\def\vl{{VL}}
\def\vlfull{{Vision Language}}
\def\zsfull{{zero-shot}}

\def\oursfull{{MAtch, eXpand and Improve}}

\NewDocumentCommand{\leonid}{ mO{} }{\noindent\textcolor{blue}{\textsf{\small[#1]}\textsuperscript{\textit{Leonid}}}}

\iccvfinalcopy % *** Uncomment this line for the final submission
\def\iccvPaperID{3246} % *** Enter the ICCV Paper ID here

\ificcvfinal\fi

\begin{document}

%%%%%%%%% TITLE
% \title{Verb Injection, Transfer learning unlabeled videos, zero-shot action recognition}
% \title{\oursfull{}: finetuning on unlabeled video collection for zero-shot and few-shot action recognition leveraging language knowledge }
% \wei{Match, Expand and Improve? Term "Unpaired"- need to keep in mind that BLIP verbs are paired}
% }
\title{\oursfull{}: Unsupervised Finetuning for Zero-Shot Action Recognition with Language Knowledge 
% \wei{Match, Expand and Improve? Term "Unpaired"- need to keep in mind that BLIP verbs are paired}
}
% Taming CLIP with verbs: zero-shot ... 

% \title{Finetuning action recognition via unpaired class dictionaries}
% \title{Finetuning VL models for action recognition via unpaired class dictionaries}
% \title{Finetuning VL models for action classification via unpaired class dictionaries}

% \and

\author{
Wei Lin$^{\dagger 1}$ \and
Leonid Karlinsky$^{2}$ \and 
Nina Shvetsova$^3$ \and 
Horst Possegger$^1$ \and 
Mateusz Kozinski$^1$ \and 
Rameswar Panda$^2$ \and
Rogerio Feris$^2$ \and
Hilde Kuehne$^{2,3}$ \and
Horst Bischof$^{1}$\and\\
$^1$Institute of Computer Graphics and Vision, Graz University of Technology, Austria\\
$^2$MIT-IBM Watson AI Lab, USA\\
$^3$Goethe University Frankfurt, Germany\\
}

\maketitle
\blfootnote{
% $\ast$ Equally contributing authors.\\ 
{
%\indent\indent\phantom{.} 
$\dagger$ Correspondence: \tt\small{wei.lin@icg.tugraz.at}}
}
\ificcvfinal\thispagestyle{empty}\fi

%%%%%%%%% ABSTRACT
\begin{abstract}
% \wei{For co-authors: Careful, abstrac and introduction (current version is from some time ago) needs to be re-written to be coherent with the method section. Method section is already pretty complete. }
    Large scale \vlfull{} (\vl{}) models have shown tremendous success in aligning representations between visual and text modalities. This enables remarkable progress in \zsfull{} recognition, image generation \& editing, and many other exciting tasks. 
    %However, these models still have some fundamental weaknesses requiring additional training before applying them to \zsfull{} action recognition tasks. 
    However, \vl{} models tend to over-represent objects while paying much less attention to verbs, and require additional tuning on video data for best \zsfull{} action recognition performance.
    While previous work relied on large-scale, fully-annotated data, in this work we propose an unsupervised approach. We adapt a \vl{} model for zero-shot and few-shot action recognition using a collection of unlabeled videos and an unpaired action dictionary. %, Large Language Models and \vl{} models, for matching, text expansion and captioning. 
    Based on that, we leverage Large Language Models and \vl{} models to build a text bag for each unlabeled video via matching, text expansion and captioning. We use those bags in a Multiple Instance Learning setup to adapt an image-text backbone to video data. 
    % We attain improvement of up to XXX\% in \zs{} transfer to novel action recognition tasks, in some cases even improving upon supervised baselines by up to YYY\%.
    Although finetuned on unlabeled video data, our resulting models demonstrate high transferability to numerous unseen zero-shot downstream tasks, improving the base \vl{} model performance by up to 14\%, and even comparing favorably to fully-supervised baselines in both zero-shot and few-shot video recognition transfer. 
    The code will be released later at \url{https://github.com/wlin-at/MAXI}.
    
   % The ABSTRACT is to be in fully-justified italicized text, at the top
   % of the left-hand column, below the author and affiliation
   % information. Use the word ``Abstract'' as the title, in 12-point
   % Times, boldface type, centered relative to the column, initially
   % capitalized. The abstract is to be in 10-point, single-spaced type.
   % Leave two blank lines after the Abstract, then begin the main text.
   % Look at previous ICCV abstracts to get a feel for style and length.
\end{abstract}

%%%%%%%%% BODY TEXT

\begin{figure}
\includegraphics[width=\columnwidth]{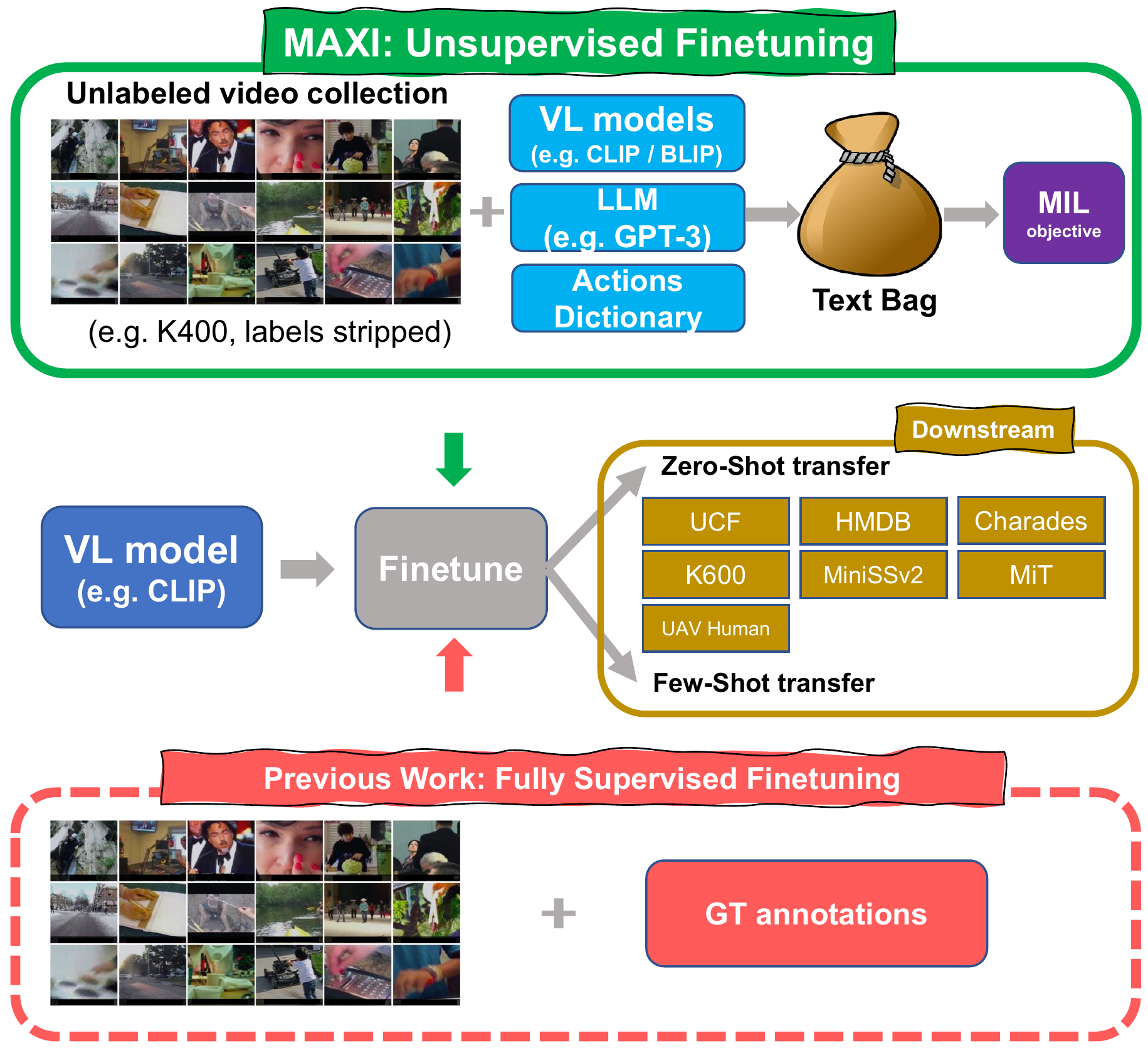}
% \vspace{-3mm}
\caption{
\label{fig:teaser}
While previous work relied on full annotation of action datasets which is time-consuming and cost-intensive to collect, our approach \OurMethod finetunes the \vl{} model with unlabeled video data. Specifically, we leverage a set of language sources (action dictionary, VL model and LLM) to construct a text bag for each unlabeled video, and employ the Multiple Instance Learning (MIL) objective for finetuning. \OurMethod demonstrates outstanding improvement of zero-shot and few-shot transfer on downstream novel action datasets. 
% \leonid{let's switch unpaired to unsupervised? or alternatively: "previous work: fully supervised finetuning" and "this work: finetuning on unlabeled video collection". Need to change in the graphic. Also, the small grey arrow coming down is not clear}
}
\end{figure}

\section{Introduction}
% \wei{For co-authors: Careful, introduction needs to be re-written to be coherent with the method section. Method section is already pretty complete. }
% \wei{Some bullet points for the introduction flow: \\
% - People have done Vl adaptation for video, but only with fully supervised data (annotated clips\\
% - People have done VL finetuning from unpaired data\\
% - we proposed to combine those two ideas and propose the first approach that does finetune a VL model from unpaired videos and given class information without any connection between the two \\
% - the scenario in this case work as follows: During training, we're given a set of unlabeled videos as well as a target dictionary with possible labels for the video set\\
% - and then describe method ... text bag to augment class labels and improve unpaired data matching\\
% }

 \vlfull{} (\vl{}) models \cite{clip, blip, align} have met unprecedented success in unlocking many vision applications \cite{clip} to work with potentially unlimited open vocabularies, through the promise of \zsfull{} transfer \cite{zhang2021tip,zhou2022conditional,zhou2022learning,gu2021open,rasheed2022bridging,zhou2022detecting,li2022language,rao2022denseclip}. This is empowered by the alignment between visual and language representation spaces, which is effectively attained by \vl{} models leveraging huge amounts of paired image and text data. Incorporating a \vl{} model as a source (base) model or as an architectural component has allowed scaling finetuning on relatively small datasets (e.g. limited in terms of the number of observed objects or other visual concepts compared to the vast \vl{} pretraining) towards \zsfull{} transfer at inference time. Such zero-shot transfer includes recognizing \cite{zhang2021tip,zhou2022conditional,zhou2022learning}, detecting \cite{gu2021open,rasheed2022bridging,zhou2022detecting}, segmenting \cite{li2022language,rao2022denseclip}, and even generating \cite{dreambooth} objects unseen during the finetuning stage and only encountered for the first time at the inference stage. 

 % However, despite all this progress, when applied directly to video data \vl{} models have been observed to suffer from several shortcomings \cite{a,b,c}. Being trained on enormous image + text paired datasets commonly makes them somewhat more vulnerable to video artifacts such as video compression and motion blur. More importantly, as has been extensively studied in several works \cite{winoground,vl-checklist,aro-benchamrk} the \vl{} models have a tendency to mostly represent the objects (nouns) appearing on the images (or video frames) resulting in a kind of `bag of object' representation while paying much less attention to visual details that would correspond to attributes (adjectives) or actions (verbs) in the language counterpart. Intuitively, both of these factors negatively affect direct \vl{} model applications to \zsfull{} action recognition resulting in significant performance drops compared to common image-based more `object-oriented' tasks. 
\vspace{4mm}

However, despite the progress in \zsfull{} image tasks, \vl{} models have been observed to underperform  when applied to \zsfull{} action recognition on video data without any finetuning \cite{wang2021actionclip,ni2022expanding,ju2022prompting,wu2023revisiting,castro2022fitclip,rasheed2022fine}. 
% Being trained on enormous image + text paired datasets commonly makes them somewhat more vulnerable to video artifacts such as video compression and motion blur. More importantly, 
% A possible reason for this might be that,
% as has been extensively studied in several works \cite{winoground,vlc,aro}, the \vl{} models have a tendency to mostly represent objects (nouns) and not actions (verbs or verb phrases).
A possible reason, as extensively studied in several works \cite{winoground,vlc,aro,hendricks2021probing}, is that \vl{} models have a tendency to mostly represent objects (nouns) and not actions (verbs or verb phrases).
Therefore, to deal with these shortcomings of \vl{} models w.r.t.~\zsfull{} action recognition, previous works \cite{wang2021actionclip,ni2022expanding,ju2022prompting,wu2023revisiting,castro2022fitclip,rasheed2022fine} have used datasets with full annotation (e.g. K400 \cite{kay2017kinetics}) to finetune \vl{} models (e.g. the most popular CLIP~\cite{clip}) towards improved video \zsfull{} recognition performance. The potential downsides of this approach are: (i) reliance on full annotation of large-scale action datasets that is time-consuming and cost-intensive, and (ii) the exposure of the model to only the limited action vocabulary during the supervised finetuning (e.g. 400 actions of K400 vs. over 8K possible single verb actions and much more possible general actions in English language) limiting the performance of \zsfull{} transfer to unseen action categories. 
% So the questions we ask in this work are: (i) is it possible to tune strong large-scale pre-trained \vl{} models towards effective \zs{} transfer for action recognition while only leveraging unlabeled video data? (ii) is it possible to harness the knowledge of language to go beyond the limited action vocabulary available for action datasets?
%
 % In our work, we answer both questions in the affirmative. 
In this context, we propose `\oursfull{}' (\OurMethod) -- to allow finetuning on completely unlabeled video data (e.g. unlabeled K400 \cite{kay2017kinetics}) and a set of language sources, such as unpaired action dictionaries, Large Language Models (LLM) (e.g. GPT-3 \cite{brown2020language}), and \vl{} models for matching (e.g. CLIP \cite{clip}) and captioning (e.g. BLIP \cite{blip}). 
To this end, \OurMethod relies on individual bags of potential texts, collected and refined based on the different language sources, that correspond to each video in the unlabeled set. It then applies Multiple Instance Learning (MIL) for finetuning the \vl{} model using those bags as illustrated in Figure \ref{fig:teaser}.
%
%As illustrated in Figure \ref{fig:teaser}, \OurMethod effectively leverages the different language sources for collecting and refining noisy bags of potential texts that may correspond to each video in the unlabeled set. It then applies Multiple Instance Learning (MIL) for finetuning the \vl{} model using those bags.
%
We extensively evaluate \OurMethod on seven downstream \zsfull{} and few-shot transfer action recognition benchmarks completely unseen during training. We show that \OurMethod is effective in leveraging unlabeled video data, not only significantly (up to 14\%) improving the source \vl{} model performance on all of those tasks, but also favorably competing with state-of-the-art supervised methods trained on fully supervised counterparts of the same finetuning data, and even improving upon them in some zero-shot and few-shot action recognition transfer tasks.
% \zsfull{} and few-shot transfer action recognition performance to video benchmarks unseen during training. 
%
% Moreover, we show that to achieve this, we can effectively fuse unlabeled video data with `bags' of language knowledge mined from either vision-conditioned and/or pure language-conditioned sources. Specifically, for vision-conditioned, we match video representations to (large) action (language) vocabulary using CLIP \cite{clip} and/or produce a sequence of frame captions using BLIP \cite{blip}. For language-conditioned, we query an LLM (e.g. GPT 3.5 \cite{gpt3.5}) for possible expanded descriptions of texts obtained in vision-conditioned ways. Applying, Multiple Instance Learning (MIL) strategies (e.g. the recent MIL-NCE \cite{milNCE}) using these `language knowledge bags' as pseudo-labels for the unlabeled videos results in the aforementioned gains.

Our contributions are as follows: (i) we propose \OurMethod{}, an approach that leverages an unlabeled video collection and a set of language sources to improve downstream \zsfull{}  action recognition; (ii) we propose to match each unlabeled video with \textit{text bags} of knowledge mined from the language sources,
 % \mk{this seems not to be true - maybe just substitute `the corresponding text bag' with `text bags'}
 and employ Multiple Instance Learning for finetuning a \vl{} model using these text bags; (iii) we extensively evaluate our approach on seven unseen action recognition benchmarks, and demonstrate up to 14\% absolute \zsfull{} performance improvements over the source \vl{} model, and even outperform baseline models trained in a fully supervised manner on the same data.
%
% To summarize, the contributions of this work are as follows: (i) we propose a new task of leveraging unlabled video collection and a set of language sources to improve downstream \zsfull{} and few-shot transfer action recognition performance for large-scale pre-trained \vl{} models; (ii) we propose \OurMethod{} -- a first approach for this task that effectively matches each unlabeled video with \textit{text bags} of knowledge mined from the language sources,
 % \mk{this seems not to be true - maybe just substitute `the corresponding text bag' with `text bags'}
 % and employs Multiple Instance Learning for finetuning a \vl{} model using these text bags; (iii) we extensively evaluate our approach on seven action recognition benchmarks unseen during training, and demonstrate up to 14\% absolute \zsfull{} performance improvements over the source \vl{} model, as well as substantial outperformance over baseline models trained in a fully supervised manner on the same data; (iv) additionally, we show a significant improvement (up to 2.8\%) in the few-shot action recognition transfer scenarios, improving upon fully supervised baseline in 10 out of 12 standard evaluations; and (v) we provide extensive ablations to validate our approach from multiple perspectives.

\begin{figure*}[ht]
\includegraphics[width=\columnwidth]{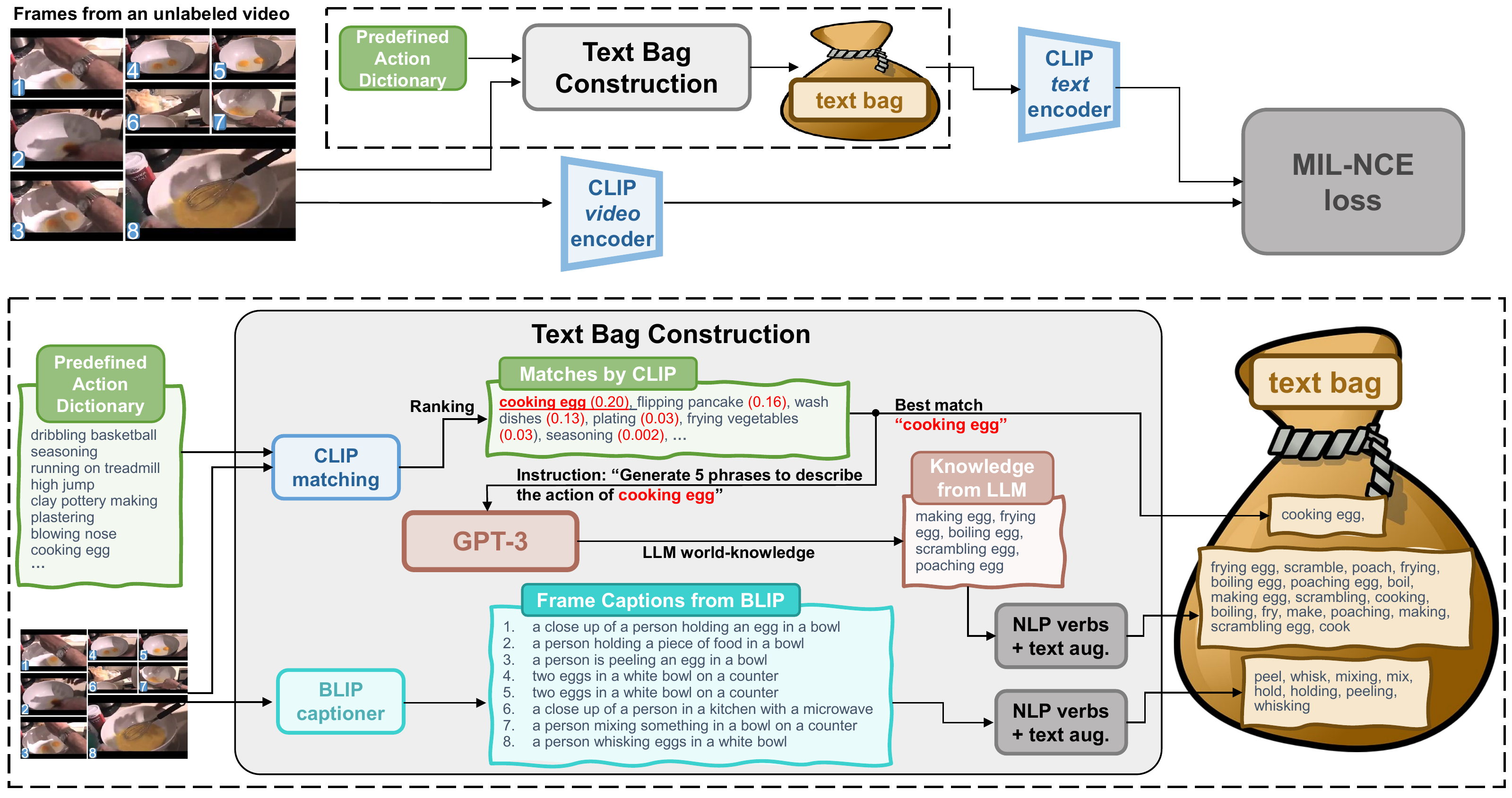}
\caption{
\label{fig:pipeline}
Pipeline of \OurMethod. Given an unlabeled video collection and a predefined action dictionary, we construct a text bag for each video. 
We finetune CLIP by passing the video and text bag through the adapted CLIP video encoder (Sec.~\ref{sec:clip_on_video}) and CLIP text encoder, and optimizing with the Multiple-Instance Learning objective (Sec. \ref{sec:mil-nce}). The text bag construction (Sec.~\ref{sec:text_bag_construction}) for an unlabeled video consists of (1) CLIP matching (2) GPT-3 text expansion and (3) BLIP captioning for video to text expansion.
}
\end{figure*}

\section{Related Work}

\noindent\textbf{Vision-language (VL) Models} revolution started with CLIP~\cite{clip} and ALIGN~\cite{align} which demonstrated that very large scale (in hundreds of millions) pre-training, on a dataset with massive amount of noisy image-text pairs collected from the web, leads to significant advances in many diverse downstream zero-shot tasks. VL models optimize for image-text alignment via contrastive learning objectives. Earlier methods, such as \cite{tan2019lxmert,chen2020uniter,li2020oscar}, relied on pre-trained object detectors to extract region features. To relax this limitation, cross-attention layers with self-supervised learning objectives, image-text matching, and masked/autoregressive language modeling were proposed in \cite{kim2021vilt,align,yang2022vision,blip}. 
%\subsection{Vision and Language (VL) Models}
BLIP~\cite{blip} combined several techniques for multi-task VL pre-training, achieving strong results in several downstream VL tasks, such as image retrieval, visual question answering~(VQA), image captioning, and reasoning tasks. Finer-level text-image alignment was attempted in~\cite{cyclip,yao2021filip,furst2021cloob,declip,gao2022pyramidclip}, employing additional losses and logic on top of the base contrastive loss of CLIP. FILIP focuses on fine-grained contrastive learning, maximizing the token-wise similarity between image and text tokens. CyClip~\cite{cyclip} employs geometrical consistency between the image and text embeddings. DeCLIP~\cite{declip} retrieves nearest neighbors for expanding the set of positive contrastive matches. While these methods have strong zero-shot results on many image benchmarks, such as ImageNet~\cite{russakovsky2015imagenet} and  MS-COCO~\cite{lin2014microsoft}, recent studies such as VL-CheckList~\cite{vlc}, the Winoground Challenge~\cite{winoground} and ARO~\cite{aro}, show that these models cannot well distinguish fine-grained language details or understand more structured concepts such as actions that commonly require understanding temporal concepts, movement, and relations between objects. In this paper, we show how \vl{} models can be adapted to better understand actions given unlabeled video data.

\myparagraph{Zero-shot action recognition} is the task of recognizing actions that have not been seen during training. This requires the bridging between visual features and semantic representations. Previous works use manually defined attributes~\cite{liu2011recognizing,zellers2017zero}, and word embeddings of action names~\cite{brattoli2020rethinking,mandal2019out,qin2017zero,shao2020temporal} or action descriptions~\cite{chen2021elaborative,qian2022rethinking,wang2017alternative} as the semantic representation. 
ER-ZSAR~\cite{chen2021elaborative} and JigsawNet~\cite{qian2022rethinking} leverage crawled descriptions of action classes with manual correction, which require efforts of human annotators for modifying the descriptions. The class descriptions are assigned to the videos based on ground truth labels. On the contrary, our text bag construction requires neither manual correction efforts nor ground truth annotation of videos. 

Recent work contributes to adapting large-scale VL model for video understanding, including zero-shot action recognition tasks~\cite{wang2021actionclip,ni2022expanding,ju2022prompting,wu2023revisiting,castro2022fitclip,rasheed2022fine}. ActionCLIP~\cite{wang2021actionclip}, Ju \etal~\cite{ju2022prompting} and XCLIP~\cite{ni2022expanding} adapt CLIP for video data with additional components for spatio-temporal modeling, and demonstrate performance improvements on video tasks. The most recent ViFi-CLIP~\cite{rasheed2022fine} shows that frame-level processing with feature pooling achieves better visual-language alignment, and outperforms sophisticated related approaches with additional learnable spatio-temporal components. In this work, we follow the architecture and finetuning paradigm of ViFi-CLIP. 

Despite the various contributions in architecture design and optimization, the related approaches still rely on ground truth annotations in finetuning CLIP for zero-shot action recognition tasks. Furthermore, no additional language source other than simple action names is explored during finetuning. \OurMethod overcomes these two limitations by finetuning CLIP (1) without any ground truth labels, and (2) expanding action names by LLM text expansion and visual captioning.

\section{Method}
In this work, we propose an approach that effectively leverages a collection of unlabeled videos and a predefined action dictionary (a potentially noisy collection of possible action text labels) to finetune the CLIP model without any ground truth annotations. The purpose of finetuning is to adapt CLIP to video data and to facilitate subsequent Zero-Shot (ZS) transfer to video recognition tasks on novel video categories which are not seen during training. We denote the predefined action dictionary as $D$, and the unlabeled video collection as $V = \{x_j | j\in I \}$, with an index set $I=\{1,...,N_V\}$. 

Our pipeline is illustrated in Fig.~\ref{fig:pipeline}. We first adapt the CLIP image encoder to a video encoder for deployment on video data (Sec.~\ref{sec:clip_on_video}). Second, given the unlabeled video collection $V$ and a predefined action dictionary $D$, we use different language sources to construct a \textit{text bag} for each video (Sec.~\ref{sec:text_bag_construction}). The text bag is a (noisy) collection of texts that potentially correspond to the video contents. Third, we perform Multiple Instance Learning (MIL) to learn from the unlabeled videos and noisy text bags (Sec. \ref{sec:mil-nce}), which allows to robustly finetune CLIP in an unsupervised manner. 
%%%%%%%%%%%%%%%%%%%%%%%%%%%%%%%%%%%%%%%%%%%%%%%% 
%%%%%%%%%%%%%%%% CLIP on Video Data
%%%%%%%%%%%%%%%%%%%%%%%%%%%%%%%%%%%%%%%%%%%%%%%%
\subsection{CLIP on Video Data}\label{sec:clip_on_video}
CLIP~\cite{clip} consists of a visual encoder $\phi_v(\cdot;\theta_v)$ and a text encoder $\phi_t(\cdot;\theta_t)$. 
We aim to adapt the CLIP image encoder for processing videos.
It is demonstrated in~\cite{rasheed2022fine} that frame-level processing on CLIP image encoder with feature pooling helps in implicitly modeling the temporal cues. This also leads to improved performance over related approaches that additionally incorporate learnable spatio-temporal components.
% leads to outperformance of sophisticated related approaches with additional learnable spatio-temporal components.
Therefore, following~\cite{rasheed2022fine}, given a video $x$, we pass $M$ frames into the visual encoder and compute the average of frame features as the video representation, \ie $z_{v} = \sum_m \phi_v(x^F_{m};\theta_v) / M $. 
An advantage of this paradigm is that the network can be initialized directly from a large-scale pretrained VL model (e.g. CLIP pretrained on 400M web image-text pairs~\cite{clip}) without adding any randomly initialized parameters. This provides a good starting point with reasonable initial performance before finetuning. We also explore extending a non-randomly-initialized-parameters paradigm to include, e.g., a parameter-free temporal-aware module (see supplementary), confirming \cite{rasheed2022fine} that a sophisticated temporal module does not lead to better video adaptation from CLIP. 
% \wei{as ViFi-CLIP~\cite{rasheed2022fine} already outperforms related work with additional learnable components, maybe we do not have to explore this in supp anymore. need to discuss. }

During inference, given a set of class prompts $C=\{t_c |^{N_C}_{c=1}\}$, the text feature is computed as $z_{t_c} = \phi_t(t_c; \theta_t )$. For simplicity, we denote the L2-normalized video feature and text feature as $z_v=\bar \phi_v(x)$ and $z_t=\bar  \phi_t(t)$.
The zero-shot classification is performed by selecting the class prompt with the maximum similarity to the video representation, \ie, $\hat c= \argmax_c \bar \phi_v(x)^\top \bar \phi_t(t_c)$. 

% zero-shot inference for pseudo labels. Knowledge distillation via pseudo label guidance. 
\subsection{Text Bag Construction}\label{sec:text_bag_construction}
Given an unlabeled video collection $V$ and a predefined action dictionary $D$ (where each item is a short sentence or a verb phrase describing an action, see Fig.~\ref{fig:pipeline}), we construct a text bag $T_i$ for each video $x_i \in V$, \ie a noisy collection of text prompts describing the video contents. 

\myparagraph{Predefined action dictionary.} In a practical scenario, we usually expect to have coarse prior knowledge of the potential action types in an unannotated video collection. The prior knowledge defines the action dictionary. To have a reasonable action dictionary, we include category names of the action dataset we use for finetuning CLIP. 
However, the prior knowledge we could obtain in a practical case might not be completely accurate. Therefore, we also explore two cases of noisy action dictionary: a) an under-specified dictionary comprised of only part of possible actions in the set, and b) an over-specified dictionary - adding noisy verbs and verb phrases randomly collected from another text corpus. An evaluation of these settings is given in Sec.~\ref{sec:robustness_noisy_dict}.
% \wei{Should we describe how the case in a practical scneario is? That we have some prior knowledge of the action dictionary for an unlabeled video collection. }

\myparagraph{CLIP matching.} For a video $x_i\in V$, we use the original CLIP to match $x_i$ with texts in $D$ w.r.t the cosine similarity. We denote the Top-1 matched text as 
\begin{equation}
% \small
\hat t_i= \argmax_{t\in D} \text{sim}(\phi_v(x_i), \phi_t(t))
\end{equation}
where $\text{sim}(u, v)=u^\mathsf{T}v / (\lVert u \rVert \lVert v \rVert)$ is the cosine similarity. We include $\hat t_i$ in the text bag $T_i$.

The CLIP matching is a means of distilling knowledge from the original CLIP as the teacher. Common choices of unlabeled video collection $V$ are usually of much smaller scale than the original CLIP domain and might be prone to overfitting. Using knowledge from the original CLIP prevents the model from overfitting to the smaller domain $V$, preserving the generalizability learned in the pretraining stage of CLIP. This hypothesis is supported by experiments in Sec.~\ref{sec:zero_shot_action_recognition} and Sec.~\ref{sec:few_shot_action_recog}, where we show that compared to all supervised finetuning baselines, the proposed unsupervised pretraining significantly improves zero-shot transfer as well as few-shot adaptation to other novel datasets. 

\myparagraph{GPT-3 text expansion.} We expand the text bag by leveraging the large-scale language model (LLM) GPT-3~\cite{brown2020language}. 
% This is similar to in-context learning \cite{} (\wei{is this correct? we do not have output, only instructions}) where few examples of inputs and desired outputs are provided in the context window of the query. 
We build upon the fact that GPT-3 has high performance on language instruction tasks~\cite{brown2020language}. %is instruction tuned \wei{is GPT-3 instruction tuned?}. 
By providing the best-matched text $\hat t_i$ in the instruction for LLM requiring it to describe this text using its language (world) knowledge (see instruction example in Fig.~\ref{fig:pipeline}), we obtain a collection of expanded alternative descriptions of the action. The descriptions contain details hallucinated by the LLM leveraging its collective world knowledge. 
We collect the verbs and verb phrases extracted from the generated expanded action descriptions. Furthermore, we perform text augmentation by including both the lemma and gerund (present participle) forms of the verbs. We add the collection of words to the text bag $T_i$. 
% \wei{need to further check this paragraph}.

\myparagraph{BLIP captioning for video to text expansion.} We employ the vision-language model BLIP~\cite{blip} for generating captions of individual frames on a video. Note that this image captioning model is not pretrained on any video domain. The frame captions provide instance-level descriptions that are dependent on the visual content of frames of the unlabeled videos. Similar to the case of GPT-3 text expansion, we collect verbs and verb phrases from these descriptions, and perform text augmentation (as stated above), adding the resulting texts to the text bag $T_i$. 

\myparagraph{Filtering text bags.} 
% \mk{Move it back to after `CLIP matching'?}
To improve the quality of the text bags, we set a threshold $\delta_p$ on the similarity score from CLIP matching. 
We determine $\delta_p$ such that $p\times 100\%$ of videos (or text bags) remain after thresholding. 
% For video $x_i\in V$, when the best matched text $\hat t_i$ has a similarity above the threshold, \ie $\text{sim}(\phi_v(x_i), \phi_t(\hat t_i)) \geq \delta_p$, we keep the text bag $T_i$ for training. 
For video $x_i\in V$, we keep the corresponding text bag $T_i$ if the best matched text $\hat t_i$ has a similarity above the threshold, \ie $\text{sim}(\phi_v(x_i), \phi_t(\hat t_i)) \geq \delta_p$.
The filtering results in a sampled index set $I_p=\{i \,|\, \text{sim}(\phi_v(x_i), \phi_t(\hat t_i)) \geq \delta_p, \forall i\in I \}$ and video set $V_p = \{x_i \,|\, i\in I_p \}$.

\subsection{Multiple Instance Learning}\label{sec:mil-nce}
We employ Multiple Instance Learning (MIL) to learn from the unlabeled videos and noisy text bags collected above. 
The MIL-NCE loss proposed in \cite{miech2020end} combines Multiple Instance Learning and Noise Contrastive Estimation.
Following MIL-NCE, instead of enforcing the match of one specific positive text to each video, we softly associate a text bag $T_i$ with each video $x_i\in V$, in which one or multiple texts could be a positive match to the video.
% \mk{what you really do, is associate to a video $x$ a union of text bags of videos that have the same pseudo-label as $x$ --- at least that's what follows from the equation; I do not think anyone will notice this, so perhaps it's safe to ignore this comment.}
As different videos have varying numbers of texts in bag, we randomly sample $N_{\text{bag}}$ texts from the original bag in each training iteration.  %$N_{\text{bag}}$
We refine the definition of the sampled text bag $T_i$ as $T_i=\{t_{i,n} |^{N_{\text{bag}}}_{n=1}\}$, where $N_{\text{bag}}$ is the constant bag size. 

The original MIL-NCE loss encourages the instance-level match between each video and its corresponding text bag. In this work, we further propose to encourage the videos and text bags, which have the same best matched text, to be close to each other. 
% (we remind that each video $x_i$ has a best matched text $\hat t_i$ from the dictionary from CLIP matching).
Noting that each video $x_i$ has a best matched text $\hat t_i$ in the dictionary from CLIP matching step, than
% \mk{what is the connection here? For example: `but we use it differently: ... '}
 our proposed loss is 
\begin{equation}
\small
    \mathcal{L}= 
    -\frac{1}{|I_B|} \sum_{i} 
    \log 
    \frac{\sum_{j} \sum_{n} \exp(\bar \phi_v(x_i)^\top \bar \phi_t (t_{j,n})/\sigma)\cdot \mathbbm{1}(\hat t_i = \hat t_j)}
    {\sum_k\sum_{n} \exp(\bar \phi_v(x_i)^\top \bar \phi_t (t_{k,n})/\sigma)}
    % -l(i_j) \cdot \log (C^I(E(i_j; \theta^I_E) ; \theta^I_C ))
\end{equation}
where $i,j,k\in I_B$ and $n\in\{1,...,N_{\text{bag}}\}$. $I_B\subset I_p$ is a sampled batch of indices. 
$t_{j,n}\in T_j$ is text in a text bag, and $\sigma$ is a temperature parameter for contrastive learning. 
$\mathbbm{1}(\hat t_i = \hat t_j)$ is an indicator that $x_i$ and $x_j$ have the same best matched text.
% \mk{it would be nice to highlight the differences between what you do and MIL-NCE in this section. I think there is enough difference to claim a contribution.}

\section{Experiments}

\subsection{Datasets}
We perform the self-supervised finetuning on Kinetics 400 (K400)~\cite{kay2017kinetics} without any ground truth labels. K400 is the most popular benchmark for action recognition tasks, containing around 240K training videos for 400 classes. We evaluate action recognition zero-shot transfer and few-shot transfer on several benchmark datasets: UCF101~\cite{soomro2012ucf101}, HMDB51~\cite{kuehne2011hmdb}, MiniSSv2~\cite{chen2021deep} (subset of SSv2~\cite{goyal2017something}), Kinetics600 (K600)~\cite{carreira2018short}, Charades~\cite{sigurdsson2016hollywood}, UAV Human (UAV)~\cite{li2021uav}, and Moments-in-Time (MiT)~\cite{monfort2019moments}. 
UCF, HMDB and K600 are collections of online user videos, which are closer in terms of style to K400. 
% are in style of online user videos which is closer to K400. 
The remaining datasets cover larger domain shifts to K400, varying from egocentric motions (MiniSSv2), human and animal videos (MiT), drone videos with small subject in frame (UAV) and 30-second long-term home videos (Charades).  
% \wei{Datasets of different levels of object bias}
% \wei{Divide the datsets into two groups: UCF, HMDB and K600, used in related work, are closer to K400. The rest has larger domain shift ... }
% \wei{Explain the challenge in each action datsaet, SSv2 - egocentric videos,  Moments-in-time: animals, UAVHuman - Drone videos with small subject in frame,   Charades is even long-term video dataset with average length of 30s. } 
More details about datasets are given in the supplementary.
% \wei{Explain that on K600, we only take the 220 classes not included in K400.}
% Experiments on Mini-Kinetics400??
% MiniSSv2 is a subset of SSv2~\cite{goyal2017something}, (87 randomly chosen classes) from~\cite{chen2021deep}

We follow the evaluation protocol of zero-shot and few-shot action recognition from~\cite{rasheed2022fine,ni2022expanding}. We report mAP for multi-label classification on Charades and Top1/Top5 accuracy for single-label classification on the remaining datasets.  
% \leonid{perhaps mention that you replicate ZS transfer and FS transfer protocolos from ViFi-CLIP, further extending it by adding supervised FS transfer (from K400 pretrain) too}

% evaluation protocol, average over three  splits on UCF, HMDB, K600. UAV has two splits. 
% We report top1 and top5 acc. mAP for Charades (multi-label classification)

\subsection{Implementation Details}
We employ CLIP with the ViT-B/16~\cite{dosovitskiy2021image} visual encoder. We follow the full-finetuning configuration of \cite{rasheed2022fine} to finetune both the visual and text encoder. 
We consistently set the temperature $\sigma$ to 0.02. For zero-shot setting, we finetune on K400 without any ground truth labels. We use the AdamW optimizer~\cite{loshchilov2017decoupled} with an initial learning rate of $5\times10^{-6}$ and a cosine decay scheduler. We sample 16 frames from each video and train with a batch size of 256 for 10 epochs. For few-shot learning, we sample 32 frames per video. We set the learning rate to $2\times10^{-6}$, and train with a batch size of 64 for 50 epochs. During inference, we sample 1 view from each video. Inspired by~\cite{wortsman2022robust, ilharco2022patching}, we perform linear weight-space ensembling between the original CLIP (with ratio of 0.2) and the finetuned model. In the main results, we set the text bag filtering ratio $p$ to 90\% and bag size to 16.

\subsection{Zero-Shot Action Recognition}\label{sec:zero_shot_action_recognition}
\begin{table*}[!tb]%[!htbp]
\scriptsize
\centering
% \begin{tabular}{@{}c M{0.8cm} M{0.8cm} M{0.8cm} M{0.7cm} M{0.7cm}M{0.7cm}@{}}
% \begin{tabular}{@{}l @{\hspace{2em}}cccccccccccccc}
\begin{tabular}{M{2.5cm}M{0.2cm}M{4cm}ccccccc}
\toprule
% Model & \multicolumn{3}{c}{TANet} & \multicolumn{3}{c}{Swin} \\
% \cmidrule(r){2-4}\cmidrule{5-7}
Method & gt & language & vis.encoder & frames &  UCF101 & HMDB51 & K600 Top1 & K600 Top5 \\
\midrule
% \midrule
% clean &  &  &  &   &  &  & & & & & & &\\
ER-ZSAR~\cite{chen2021elaborative} & yes & Manual description & TSM & 16 &  51.8 $\pm$ 2.9 & 35.3 $\pm$ 4.6 & 42.1 $\pm$ 1.4 & 73.1 $\pm$ 0.3\\
JigsawNet~\cite{qian2022rethinking} & yes & Manual description & R(2+1)D & 16 & 56.0 $\pm$ 3.1 & 38.7 $\pm$ 3.7 & - & -  \\
ActionCLIP~\cite{wang2021actionclip} & yes & K400 dict. & ViT-B/16 & 32  &  58.3 $\pm$ 3.4 & 40.8 $\pm$ 5.4 & 66.7 $\pm$ 1.1 & 91.6 $\pm$ 0.3 \\
XCLIP~\cite{ni2022expanding} & yes & K400 dict.  & ViT-B/16  & 32 &   72.0 $\pm$ 2.3 & 44.6 $\pm$ 5.2 & 65.2 $\pm$ 0.4 & 86.1 $\pm$ 0.8 \\
A5~\cite{ju2022prompting} & yes & K400 dict. & ViT-B/16  & 32  &    69.3 $\pm$ 4.2 & 44.3 $\pm$ 2.2 & 55.8 $\pm$ 0.7 & 81.4 $\pm$ 0.3 \\
ViFi-CLIP~\cite{rasheed2022fine}* & yes & K400 dict. & ViT-B/16 & 16 &    74.9 $\pm$ 0.6 & 50.9 $\pm$ 0.7 & 67.7 $\pm$ 1.1 & 90.8 $\pm$ 0.3  \\
ViFi-CLIP~\cite{rasheed2022fine} & yes & K400 dict. & ViT-B/16 & 32 &   76.8 $\pm$ 0.7 & 51.3 $\pm$ 0.6 & 71.2 $\pm$ 1.0 & 92.2 $\pm$ 0.3 \\
Text4Vis~\cite{wu2023revisiting} & yes & K400 dict. & ViT-L/14 & 16 &  - & - & 68.9 $\pm$ 1.0 & -\\
% Open-VCLIP~\cite{weng2023transforming} \wei{to remove} & yes & ViT-B/16 & 8 & 3 & 83.4 $\pm$ 1.2 & 53.9 $\pm$ 1.2 & 73.0 $\pm$ 0.8 & 93.2 $\pm$ 0.1 \\
\midrule
CLIP~\cite{clip} & no & - & ViT-B/16 & 16 &  69.9 $\pm$ 1.3 & 38.0 $\pm$ 1.7 & 63.5 $\pm$ 0.4 & 86.8 $\pm$ 0.4   \\
\OurMethod  & no & K400 dict. & ViT-B/16 & 16 & 76.6 $\pm$ 0.9 & 50.5 $\pm$ 0.9 & 70.4 $\pm$ 0.8 & 91.5 $\pm$ 0.3\\
\OurMethod  & no & K400 dict, GPT3 verbs & ViT-B/16 & 16 & \underline{77.8} $\pm$ 0.3 & 51.6 $\pm$ 0.9 & \textbf{71.6} $\pm$ 1.0 & 92.3 $\pm$ 0.3  \\
\OurMethod & no & K400 dict, GPT3 verbs & ViT-B/16 & 16/32 &  \underline{77.8} $\pm$ 0.5 & 51.9 $\pm$ 1.1 & \textbf{71.6} $\pm$ 1.0 & \underline{92.4}  $\pm$ 0.3 \\
\OurMethod & no & K400 dict, GPT3 verbs, BLIP verbs & ViT-B/16 & 16 &  \textbf{78.2} $\pm$ 0.8  & \underline{52.2} $\pm$ 0.6 & 71.4 $\pm$ 0.9 & \textbf{92.5}  $\pm$ 0.3    \\
\OurMethod  & no & K400 dict, GPT3 verbs, BLIP verbs & ViT-B/16 & 16/32 &  \textbf{78.2} $\pm$ 0.8 & \textbf{52.3} $\pm$ 0.7 & \underline{71.5} $\pm$ 0.8 & \textbf{92.5} $\pm$ 0.4     \\

\bottomrule
\end{tabular}
\caption{
Zero-shot action recognition on UCF101, HMDB51 and K600. We report mean and standard deviation of results on three official validation splits. All models (except for the original CLIP) are trained on K400. We set the text bag filtering ratio $p$ to 90\%. We train with 16 frames per video and report single-view inference results with 16 and 32 frames here. 
*denotes our re-evaluation.   
% \wei{best numbers in bold, second best in blue color, and add improvement}.
 }
\label{tab:zs_action_recog_sota}
% \label{tab:yti_state_of_the_art}
\end{table*}
\begin{table*}[!tb]%[!htbp]
% \footnotesize
\scriptsize
\centering
% \begin{tabular}{@{}c M{0.8cm} M{0.8cm} M{0.8cm} M{0.7cm} M{0.7cm}M{0.7cm}@{}}
% \begin{tabular}{@{}l @{\hspace{2em}}cccccccccccccc}
\begin{tabular}{cccccccc}
\toprule
% Model & \multicolumn{3}{c}{TANet} & \multicolumn{3}{c}{Swin} \\
% \cmidrule(r){2-4}\cmidrule{5-7}
Method & gt & language & Charades & MiT & MiniSSv2 & UAV \\
\midrule

% \midrule
% clean &  &  &  &   &  &  & & & & & & &\\

ViFi-CLIP~\cite{rasheed2022fine} & yes  & K400 dict.  &  \textbf{25.77} & 21.68 / 44.19 & 5.98 / \textbf{19.04} & \textbf{4.67} / \textbf{15.18}   \\
\midrule
CLIP~\cite{clip} & no & - & 19.80 & 20.11 / 40.81 & 3.96 / 14.42 & 1.79 / 7.05    \\
% \midrule
\OurMethod & no & K400 dict. &  23.47 & 21.94 / 45.68 & 5.19 / 17.71 & 2.42 / 8.39\\
\OurMethod & no & K400 dict., GPT3 verbs & 23.74  & \underline{22.11} / \underline{45.79} & 5.60 / 16.73 & \underline{2.77} / \underline{9.07}  \\
\OurMethod & no & K400 dict., GPT3 verbs, BLIP verb & \underline{23.79} & \textbf{22.91} / \textbf{46.38} & \textbf{6.37} / \underline{18.73} & 2.72 / 9.00 \\

\bottomrule
\end{tabular}
\caption{
Zero-shot action recognition on Charades, MiT, MiniSSv2 and UAV. All models (except for CLIP) are trained on K400. We report the mAP of multi-label classification on Charades and Top-1/Top-5 single-label classification accuracy for MiT, MiniSSv2 and UAV. We set the text bag filtering ratio $p$ to 90\%. }
\label{tab:zs_action_recog_sota_part2}
\end{table*}

% \begin{table*}
% \scriptsize
% \floatbox[{\capbeside\thisfloatsetup{capbesideposition={right,top},capbesidewidth=5.18cm,capbesidesep=quad}}]{table}[\FBwidth]
% {\caption{Zero-shot action recognition on Charades, MiT, MiniSSv2 and UAV. All models (except for CLIP) are trained on K400. We report the mAP of multi-label classification on Charades and Top-1/Top-5 accuracy for MiT, MiniSSv2 and UAV. 
% % We set the text bag filtering ratio $p$ to 90\%.  \wei{remove  UAV Human?}
% }
% \label{tab:zs_action_recog_sota_part2}}
% {
% \begin{tabular}{M{0.8cm}M{0.8cm}M{2.5cm}M{1.2cm}M{1.3cm}M{1.2cm}M{1.2cm}}
% \toprule
% Method & gt & language & Charades & MiT & MiniSSv2 & UAV \\
% \midrule
% ViFi-CLIP~\cite{rasheed2022fine} & yes  & K400 dict.  &  \textbf{25.77} & 21.68 / 44.19 & 5.98 / \textbf{19.04} & \textbf{4.67} / \textbf{15.18}   \\
% \midrule
% CLIP~\cite{clip} & no & - & 19.80 & 20.11 / 40.81 & 3.96 / 14.42 & 1.79 / 7.05    \\
% % \midrule
% \OurMethod & no & K400 dict. &  23.47 & 21.94 / 45.68 & 5.19 / 17.71 & 2.42 / 8.39\\
% \OurMethod & no & K400 dict., GPT3 verbs & 23.74  & \underline{22.11} / \underline{45.79} & 5.60 / 16.73 & \underline{2.77} / \underline{9.07}  \\
% \OurMethod & no & K400 dict., GPT3 verbs, BLIP verb & \underline{23.79} & \textbf{22.91} / \textbf{46.38} & \textbf{6.37} / \underline{18.73} & 2.72 / 9.00 \\
% \bottomrule
% \end{tabular}
% }
% \end{table*}
We finetune CLIP on the large-scale K400 dataset stripped of the original ground truth labels. We perform zero-shot action recognition on seven different datasets to verify that cross-dataset model generalizability transfer after the finetuning. In zero-shot setting, the model is evaluated directly on downstream datasets with unseen classes, without being trained on any samples of these datasets.
% via text queries of \textit{A video of $<$action label$>$}, without being trained on any samples of these datasets.

In Table~\ref{tab:zs_action_recog_sota}, we first compare to other state-of-the-art methods, all of which use K400 to adapt CLIP models for zero-shot recognition tasks on UCF, HMDB and K600. Following~\cite{rasheed2022fine,ni2022expanding,chen2021elaborative}, we report the mean and standard deviation of results on three official validation sets. ER-ZSAR~\cite{chen2021elaborative} and JigsawNet~\cite{qian2022rethinking} are zero-shot action recognition approaches that train with K400 ground truth annotations. They leverage crawled descriptions of action classes with manual correction, which requires efforts from human annotators. Afterwards, the class descriptions are assigned to videos based on ground truth annotations.
% is a language-enhanced action recognition approach prior to CLIP, and uses separate TSM spatio-temporal visual encoder~\cite{lin2019tsm} and a BERT text encoder~\cite{devlin2018bert}. It explores language sources of Wikipedia and WordNet. 
We see that the original CLIP has good direct zero-shot performance across the three datasets, which performs better or on par with ER-ZSAR~\cite{chen2021elaborative} and JigsawNet~\cite{qian2022rethinking}. 
The rest of the compared approaches all adapt CLIP models on video-text pairs with the K400 ground truth class labels as texts. 
% The text is from the K400 action dictionary. 
Among them, the most recent ViFi-CLIP~\cite{rasheed2022fine} achieves the best result, outperforming all the other approaches, without adding any learnable spatio-temporal modules (as done by other approaches such as~\cite{wang2021actionclip,ju2022prompting,ni2022expanding}). 

In a similar full finetuning paradigm to ViFi-CLIP, \OurMethod achieves favorable results without using any ground truth annotation. We report the performance of \OurMethod with different combinations of language sources. Simply with the original K400 action dictionary, we already outperform most of the related work across the three datasets. With the additional GPT-3 verbs and BLIP verbs in the text bag, we further boost the performance, achieving the state-of-the-art among the three datasets. 

For a thorough analysis of the model generalizibility, we further report the performance of \OurMethod on four datasets (Charades, MiT, MiniSSv2 and UAV) with larger domain shift to K400 in Table~\ref{tab:zs_action_recog_sota_part2}. 
%In Table~\ref{tab:zs_action_recog_sota_part2}, for a thorough analysis of the model generalizibility, we further report the performance of \OurMethod on four datasets (Charades, MiT, MiniSSv2 and UAV) with larger domain shift to K400. 
In comparison to the original CLIP, our finetuned model has improved zero-shot transfer on all datasets. With the additional language sources of GPT-3 and BLIP, we even outperform ViFi-CLIP trained with ground truth of K400, on the challenging MiT and MiniSSv2 datasets. 

\subsection{Few-Shot Action Recognition}\label{sec:few_shot_action_recog}
We perform few-shot all-way action recognition to evaluate the model learning capacity in a low data regime. 
In this setting, we specifically verify whether our self-supervised finetuning on K400 provides a proper initialization for few-shot learning. 
We follow the few-shot configuration of ViFi-CLIP~\cite{rasheed2022fine} and XCLIP~\cite{ni2022expanding}, and use the same training samples in 2, 4, 8 and 16-shot experiments without additional language source for a fair comparison. We train with 32 frames per video. We use the best backbone of self-supervised finetuning (from Sec.~\ref{sec:zero_shot_action_recognition}) as the model initialization for few-shot training. In Table~\ref{tab:few_shot_table}, we report few-shot results of~\OurMethod on three datasets, and also the zero-shot performance of our initialization as a reference. We compare with related approaches that directly perform few-shot learning on CLIP. For a fair comparison, we include the result of few-shot training with a CLIP model that is pre-trained with ground truth labels in the ViFi-CLIP paradigm. 

We see that few-shot learning using a \OurMethod-pretrained backbone leads to best performance in most settings, even outperforming the fully-supervised pretrained backbone of ViFi-CLIP. The performance gap is significant in the more challenging extremely limited data scenarios (\eg 2-shot on HMDB and UCF). Pretraining with full supervision as an initialization might lead to degraded performance in the following few-shot learning (\eg 8-shot on HMDB, 4-shot on UCF), while our self-supervised finetuned model mitigates this problem, indicating improved generalizability. 

\begin{table*}[!tb]%[!htbp]
\footnotesize
\centering
% \begin{tabular}{@{}c M{0.8cm} M{0.8cm} M{0.8cm} M{0.7cm} M{0.7cm}M{0.7cm}@{}}
% \begin{tabular}{@{}l @{\hspace{2em}}cccccccccccccc}
\begin{tabular}{ccccccccccccccc}
\toprule
% Model & \multicolumn{3}{c}{TANet} & \multicolumn{3}{c}{Swin} \\
% \cmidrule(r){2-4}\cmidrule{5-7}
Dataset & \multirow{2}*{pretrain on K400} & \multirow{2}*{sett.} & \multicolumn{4}{c}{HMDB51}  & \multicolumn{4}{c}{UCF101} & \multicolumn{4}{c}{SSv2}  \\
\cmidrule(r){4-7}\cmidrule(r){8-11}\cmidrule(r){12-15}
Shots & ~ & ~ & 2 & 4 & 8 & 16 & 2 & 4 & 8 & 16 & 2 & 4 & 8 & 16 \\
\midrule
CLIP~\cite{clip} & no & ZS &  41.9 & 41.9 & 41.9 & 41.9 & 63.6 & 63.6 & 63.6 & 63.6 & 2.7 & 2.7 & 2.7 & 2.7\\
ActionCLIP~\cite{wang2021actionclip} & no & FS & 47.5 & 57.9 & 57.3 & 59.1 & 70.6 & 71.5 & 73.0 & 91.4 & 4.1 & 5.8 & 8.4 & \underline{11.1} \\
XCLIP~\cite{ni2022expanding} & no & FS & 53.0 & 57.3 & 62.8 & 64.0 & 48.5 & 75.6 & 83.7 & 91.4 & 3.9 & 4.5 & 6.8 & 10.0 \\
A5~\cite{ju2022prompting} & no & FS & 39.7 & 50.7 & 56.0 & 62.4 & 71.4 & 79.9 & 85.7 & 89.9 & 4.4 & 5.1 & 6.1 & 9.7 \\
ViFi-CLIP~\cite{rasheed2022fine} & no & FS & \underline{57.2} & \textbf{62.7} & \underline{64.5} & \textbf{66.8} & 80.7 & 85.1 & 90.0 & 92.7 & 6.2 & 7.4 & 8.5 & \textbf{12.4}   \\
\midrule
\OurMethod & yes w/o gt & ZS & 49.2 & 49.2 & 49.2 & 49.2 & 77.8 & 77.8 & 77.8 & 77.8 & 4.8 & 4.8 & 4.8 & 4.8 \\
ViFi-CLIP~\cite{rasheed2022fine} & yes gt & FS & 55.8 & \underline{60.5} & 64.3 & 65.4 & \underline{84.0} & \underline{86.5} & \underline{90.3} & \underline{92.8} & \underline{6.6} & \underline{6.8} & \underline{8.6} & 11.0 \\
\OurMethod & yes w/o gt & FS & \textbf{58.0} & 60.1 & \textbf{65.0} & \underline{66.5} & \textbf{86.8} & \textbf{89.3} & \textbf{92.4} & \textbf{93.5} & \textbf{7.1} & \textbf{8.4} & \textbf{9.3} & \textbf{12.4} \\

\bottomrule
\end{tabular}
\caption{
Few-shot action recognition on HMDB, UCF and SSv2. We report few-shot learning results with and without pretraining on K400. %It shows that the propsoed \OurMethod-pretrained backbone leads to best performance in most settings, even outperforming fully-supervised pretrained backbones.
 }
\label{tab:few_shot_table}
% \label{tab:yti_state_of_the_art}
\end{table*}

\begin{table*}[!tb]%[!htbp]
\scriptsize
\centering
% \begin{tabular}{@{}c M{0.8cm} M{0.8cm} M{0.8cm} M{0.7cm} M{0.7cm}M{0.7cm}@{}}
% \begin{tabular}{@{}l @{\hspace{2em}}cccccccccccccc}
\begin{tabular}{ccccccccccccccc}
\toprule
Matching & ratio $p$ & matching acc. on K400 & UCF101 & HMDB51 & K600 & MiniSSv2 & Charades & UAV Human & Moments-in-time \\
\midrule
\multicolumn{3}{c}{CLIP~\cite{clip} (w/o finetune) Zero-Shot} & 69.93 & 38.02 & 63.48 & 3.96 & 19.80 & 1.79 & 20.11 \\
\midrule
gt & 100\% & 100\% & \textbf{82.39} & \textbf{52.68} & \textbf{73.39} & 5.61 & \textbf{25.31} & \textbf{4.47} & \textbf{23.79} \\
\midrule
CLIP matching & 100\% & 59.7\% & 77.88 & 51.09 & 71.24 & 5.46 & 23.52 & 2.53 & 22.44  \\
CLIP matching & 90\% & 64.3\% & 78.17 & \underline{52.24} & \underline{71.43} & \textbf{6.37} & 23.79 & 2.72 & \underline{22.91} \\
CLIP matching & 50\% & 80.9\% & \underline{78.18} & 50.35 & 70.78 & \underline{5.74} & \underline{23.89} & \underline{3.06} & 22.41 \\
CLIP matching & 30\% & 89.5\% & 76.71 & 47.73 & 70.57 & 4.92 & 23.14 & 2.89 & 21.96 \\

\bottomrule
\end{tabular}
\caption{
Text bag filtering with different filtering ratio $p$. We report the CLIP matching accuracy (after filtering) on K400, and the zero-shot transfer performance of models finetuned with the filtered K400 videos and text bags. 
 }
\label{tab:ps_analysis}
% \label{tab:yti_state_of_the_art}
\end{table*}
\subsection{Ablation Study}\label{sec:ablation_study}
\subsubsection{Text bag filtering}\label{sec:text_bag_filtering}
% \myparagraph{Text bag filtering.} 
To improve the quality of text bags used in training, we set a threshold $\delta_p$ on the similarity score from CLIP matching, such that $p\times 100\%$ of videos with highest similarity scores remain after the thresholding (see Sec.~\ref{sec:text_bag_construction}). We perform CLIP matching between unlabeled K400 videos and the K400 action dictionary, and use the filtered videos and text bags for finetuning CLIP. In Table~\ref{tab:ps_analysis}, we report the matching accuracy (after filtering), and zero-shot transfer performance of models finetuned with the filtered K400 videos and text bags. As a reference, we also report CLIP zero-shot performance, and the case of finetuning on 100$\%$ accurate video-textbag pairs using ground truth annotation, which leads to the best zero-shot transfer on most datasets. 

In Table~\ref{tab:ps_analysis}, we notice that the CLIP matching accuracy increases continuously with decreasing filtering ratio $p$. Setting $p=90\%$ leads to consistent improvement of zero-shot transfer, in comparison to the case of $p=100\%$ due to improved quality of matched texts. 
Setting $p=50\%$ leads to partial improvement compared to $p=100\%$. Further reducing $p$ to $50\%$ leads to performance degradation due to the limited amount of data. This indicates that selecting text bags that CLIP is confident about ensures improved finetuning for more effective zero-shot transfer. However, there is a trade-off between the quality of the filtered data and the amount of data used for training.

% \myparagraph{Robustness of finetuning with noisy action dictionary.} 
\subsubsection{Robustness against noisy action dictionary}\label{sec:robustness_noisy_dict}
In a practical scenario, we have coarse prior knowledge of the potential action types in an unannotated video collection, which defines an action dictionary. However, such knowledge might be noisy. We explore the robustness of our finetuning pipeline against such a noisy action dictionary. We consider two cases of noisy action dictionaries: (1) an under-specified dictionary consisting of only half of the words of the original K400 action dictionary. Specifically, we use the 200 action names from MiniKinetics~\cite{chen2021deep} (a 200-class subset of K400). (2)~An over-specified dictionary by adding noisy verbs and verb phrases into the original K400 action dictionary. We parse verbs from the captions in the validation set of the WebVid2.5M dataset~\cite{bain2021frozen}, and randomly sample 400 verbs to add to the dictionary, resulting in a dictionary of 800 verbs or verb phrases. 

In Table~\ref{tab:noisy_class_space}, we report the zero-shot transfer performance of models finetuned with these noisy dictionaries. Here we set the text bag filtering $p=50\%$ for improved text bag quality. We also report the results with the original K400 action dictionary as a reference. Apparently, using the clean original K400 action dictionary leads to the best zero-shot transfer on most of the downstream datasets. However, using noisy action dictionaries still leads to significant performance boost compared to the CLIP zero-shot results without finetuning. This indicates the robustness of our pipeline with different cases of noisy predefined dictionaries. 

\begin{table*}[!tb]%[!htbp]
\scriptsize
\centering
% \begin{tabular}{@{}c M{0.8cm} M{0.8cm} M{0.8cm} M{0.7cm} M{0.7cm}M{0.7cm}@{}}
% \begin{tabular}{@{}l @{\hspace{2em}}cccccccccccccc}
\begin{tabular}{ccccccccccccccc}
\toprule
Action dictionary & dictionary size & UCF101 & HMDB51 & K600 & MiniSSv2 & Charades & UAV Human & Moments-in-time \\
\midrule
\multicolumn{2}{c}{CLIP~\cite{clip} (w/o finetune) Zero-Shot} & 69.93 / 92.7 & 38.02 / 66.34 & 63.48 / 86.80 & 3.96 / 14.42 & 19.80 & 1.79 / 7.05 & 20.11 / 40.81 \\
\midrule
K400 & 400 & \textbf{78.18} / \textbf{96.03} & \textbf{50.35} / \textbf{77.10} &  \textbf{70.78} / \textbf{92.17} & \underline{5.74} / \underline{17.70} & \textbf{23.89} & \textbf{3.06} / \textbf{9.46} & \underline{22.41} / \underline{45.83} \\
MiniKinetics & 200 & 75.10 / 95.82 & \underline{48.34} / \underline{76.95} & \underline{69.23} / 90.92 & \textbf{6.50} / \textbf{18.76} & \underline{22.70} & \underline{2.40} / \underline{8.04} & \textbf{22.50} / \textbf{46.01} \\
K400+WebVid2.5M & 800 & \underline{75.99} / \underline{96.00} & 45.97 / 73.94 & 69.14 / \underline{91.13} & 4.81 / 15.79 & 22.67 & 2.11 / 8.00 & 20.92 / 43.99 \\
% UCF101 & 101 & -  \\

\bottomrule
\end{tabular}
\vspace{-2mm}
\caption{
Robustness of finetuning with noisy action dictionaries. We report the zero-shot transfer performance (mAP on Charades and Top1/Top5 accuracy on other datasets). We set the text bag filtering ratio $p=50\%$ for improved text bag quality.
}
\label{tab:noisy_class_space}
% \label{tab:yti_state_of_the_art}
\end{table*}

% \myparagraph{What words to include in the text bag?} 
\subsubsection{What words to include in the text bag?}\label{sec:words_in_text_bag}
In Table~\ref{tab:bag_of_words}, we investigate different combinations of words to include in the text bag. Besides the original K400 action dictionary (\textit{K400 dict.}), we explore: (1) \textit{BLIP verbs}: verbs parsed from BLIP captions; (2) \textit{BLIP object nouns}: nouns of objects parsed from BLIP captions; (3) \textit{GPT3 verbs}: verbs and verb phrases from GPT3 text expansion. For a thorough ablation, we set the text bag filtering ratio $p=100\%$ to keep the full noisy text bag property.

In Table~\ref{tab:bag_of_words}, we notice that additional language source upon the original K400 action dictionary leads to further improvement in zero-shot transfer. Interestingly, using BLIP verbs has slightly better results than the case of BLIP object nouns. We assume this is because CLIP has a high object bias and is less sensitive to the language of verbs. Finetuning CLIP by injecting verbs leads to better zero-shot performance in action recognition. Consequently, combining BLIP verbs and GPT3 verbs in the text bag leads to the best zero-shot transfer.

% \begin{table*}[!tb]%[!htbp]
% \small
% \centering
% % \begin{tabular}{@{}c M{0.8cm} M{0.8cm} M{0.8cm} M{0.7cm} M{0.7cm}M{0.7cm}@{}}
% % \begin{tabular}{@{}l @{\hspace{2em}}cccccccccccccc}
% \begin{tabular}{ccccccccccccccc}
% \toprule
% % Model & \multicolumn{3}{c}{TANet} & \multicolumn{3}{c}{Swin} \\
% % \cmidrule(r){2-4}\cmidrule{5-7}
% % Dataset & \multirow{2}*{pretrain} & \multirow{2}*{sett.} & \multicolumn{4}{c}{HMDB51}  & \multicolumn{4}{c}{UCF101} & \multicolumn{4}{c}{SSv2}  \\
% % \cmidrule(r){4-7}\cmidrule(r){8-11}\cmidrule(r){12-15}
% Text bags & UCF101 & HMDB51 & K600 & MiniSSv2 & Charades & UAV Human & Moments-in-time \\
% \midrule
% predefined vocabulary & 76.45 & 47.43 &  69.98 & 4.76 & 22.62 & 2.39 & 20.72 \\
% predefined vocabulary + BLIP object nouns & 77.23 & 50.15 & 71.13 & 5.45 & 23.43 & 2.42 & 22.15 \\
% predefined vocabulary + BLIP verbs & 76.94 & 50.92 & 71.25 & 5.73 & 23.64 & 2.70 & 22.73\\
% predefined vocabulary + GPT3 verbs & 76.98 & 50.46 & 71.24 & 6.52 & 23.54 & 2.62 & 22.91\\
% predefined vocabulary + GPT3 verbs + BLIP verbs & 77.88 & 51.09 & 71.24 & 5.46 & 23.52 & 2.53 & 22.44 \\

% \bottomrule
% \end{tabular}
% \caption{
% Contents in Text bags. 100 \% pseudo labels. \wei{only show results on the first 3 datasets}.  
%  }
% \label{tab:bag_of_words}
% % \label{tab:yti_state_of_the_art}
% \end{table*}

\begin{table}[!tb]%[!htbp]
\scriptsize
\centering
% \begin{tabular}{@{}c M{0.8cm} M{0.8cm} M{0.8cm} M{0.7cm} M{0.7cm}M{0.7cm}@{}}
% \begin{tabular}{@{}l @{\hspace{2em}}cccccccccccccc}
\begin{tabular}{cccc}
\toprule
% Model & \multicolumn{3}{c}{TANet} & \multicolumn{3}{c}{Swin} \\
% \cmidrule(r){2-4}\cmidrule{5-7}
% Dataset & \multirow{2}*{pretrain} & \multirow{2}*{sett.} & \multicolumn{4}{c}{HMDB51}  & \multicolumn{4}{c}{UCF101} & \multicolumn{4}{c}{SSv2}  \\
% \cmidrule(r){4-7}\cmidrule(r){8-11}\cmidrule(r){12-15}
Text bag & UCF101 & HMDB51 & K600  \\
\midrule
K400 dict. & 76.45 & 47.43 &  69.98  \\
K400 dict. + BLIP object nouns & 76.23 & 50.15 & 71.13 \\
K400 dict. + BLIP verbs & 76.94 & \underline{50.92} & \textbf{71.25}\\
K400 dict. + GPT3 verbs & \underline{76.98} & 50.46 & \underline{71.24} \\
K400 dict. + GPT3 verbs + BLIP verbs & \textbf{77.88} & \textbf{51.09} & \underline{71.24}  \\

\bottomrule
\end{tabular}
\caption{
Combinations of words in text bags. We report the zero-shot transfer performance on UCF, HMDB and K600. For a thorough ablation, we set the text bag filtering ratio $p=100\%$ to keep the full noisy text bag property.  
 }
\label{tab:bag_of_words}
% \label{tab:yti_state_of_the_art}
\end{table}

% \myparagraph{How to learn from words in text bags?} 
\subsubsection{How to learn from words in text bags?}\label{sec:how_to_learn_from_text_bag}
In Table~\ref{tab:ablation_mil_nce}, we explore different strategies of learning from words in a text bag: (1) \textit{Cross entropy}: classification in a fixed class space. (2) \textit{NCE}: contrastive learning to encourage instance-level match between a pair of video and text. In this case, we randomly sample one text from the text bag in each iteration. (3) \textit{MIL-Max}: in each iteration, among words in a text bag, we choose the word with the maximum similarity to the video, and pass the similarity in the contrastive loss. 
(4) \textit{MIL-NCE}: as explained in Sec.~\ref{sec:mil-nce}, we softly associate a bag of texts with the video, and sum up the similarities of texts in a bag (5)  \textit{MIL-NCE only instance-level}: the \textit{MIL-NCE} on instance-level match between video and text bag, without encouraging videos and text bags with the same best matched text to be close to each other (see Sec.~\ref{sec:mil-nce}). 
In Table~\ref{tab:ablation_mil_nce}, we see that cross entropy of classification in a fixed class space leads to the most inferior result, while our MIL-NCE achieves the best improvement. Encouraging videos and text bags with the same best matched text to be close to each other also leads to some performance boost in contrast to only instance-level matching.
%upon case of only instance-level match. 

% \begin{table*}[!tb]%[!htbp]
% \small
% \centering
% % \begin{tabular}{@{}c M{0.8cm} M{0.8cm} M{0.8cm} M{0.7cm} M{0.7cm}M{0.7cm}@{}}
% % \begin{tabular}{@{}l @{\hspace{2em}}cccccccccccccc}
% \begin{tabular}{ccccccccccccccc}
% \toprule
% % Model & \multicolumn{3}{c}{TANet} & \multicolumn{3}{c}{Swin} \\
% % \cmidrule(r){2-4}\cmidrule{5-7}
% % Dataset & \multirow{2}*{pretrain} & \multirow{2}*{sett.} & \multicolumn{4}{c}{HMDB51}  & \multicolumn{4}{c}{UCF101} & \multicolumn{4}{c}{SSv2}  \\
% % \cmidrule(r){4-7}\cmidrule(r){8-11}\cmidrule(r){12-15}
% Text bags & UCF101 & HMDB51 & K600 & MiniSSv2 & Charades & UAV Human & Moments-in-time \\
% \midrule
% Cross entropy & 74.48 & 48.69 & 65.09 & 5.24 & 22.91 & 2.60 & 21.64 \\
% NCE & 77.26 & 49.85 & 70.08 & 4.75 & 23.45 & 2.62 & 21.71\\
% MIL-Max & 77.24 & 49.85 & 70.71 & 5.09 & 23.49 & 2.46 & 21.84\\
% MIL-NCE w/o label mask & 76.96 & 48.48 & 70.14 \\
% MIL-NCE & 77.88 & 51.09 & 71.24 & 5.46 & 23.52 & 2.53 & 22.44 \\
% \bottomrule
% \end{tabular}
% \caption{
% Different strategies of learning from text bags.  We use text bags of GPT3 verbs + BLIP verbs. 100\% pseudo labels. \wei{only show the first 3 datasets}.
%  }
% \label{tab:ablation_mil_nce}
% % \label{tab:yti_state_of_the_art}
% \end{table*}

\begin{table}[!tb]%[!htbp]
\footnotesize
\centering
% \begin{tabular}{@{}c M{0.8cm} M{0.8cm} M{0.8cm} M{0.7cm} M{0.7cm}M{0.7cm}@{}}
% \begin{tabular}{@{}l @{\hspace{2em}}cccccccccccccc}
\begin{tabular}{cccc}
\toprule
% Model & \multicolumn{3}{c}{TANet} & \multicolumn{3}{c}{Swin} \\
% \cmidrule(r){2-4}\cmidrule{5-7}
% Dataset & \multirow{2}*{pretrain} & \multirow{2}*{sett.} & \multicolumn{4}{c}{HMDB51}  & \multicolumn{4}{c}{UCF101} & \multicolumn{4}{c}{SSv2}  \\
% \cmidrule(r){4-7}\cmidrule(r){8-11}\cmidrule(r){12-15}
Objective & UCF101 & HMDB51 & K600 \\
\midrule
Cross entropy & 74.48 & 48.69 & 65.09  \\
NCE & \underline{77.26} & 49.85 & 70.08 \\
MIL-Max & 77.24 & 49.85 & \underline{70.71} \\
MIL-NCE only instance-level & 76.96 & \underline{50.48} & 70.14 \\
MIL-NCE & \textbf{77.88} & \textbf{51.09} & \textbf{71.24} \\

\bottomrule
\end{tabular}
\vspace{-2mm}
\caption{
Different strategies of learning from text bags. 
% We use text bags of GPT3 verbs + BLIP verbs. 
We report the zero-shot transfer performance on UCF, HMDB and K600.
For a thorough ablation, we set the text bag filtering ratio $p=100\%$ to keep the full noisy text bag property.
 }
\label{tab:ablation_mil_nce}
% \label{tab:yti_state_of_the_art}
\end{table}

% \myparagraph{Bag size.} 
\subsubsection{Bag size}\label{sec:bag_size}
We perform an ablation on the bag size in Table~\ref{tab:bag_size}. A bag size of 1 is the same as \textit{NCE} loss with random word sampling in Table~\ref{tab:ablation_mil_nce}. Increasing the bag size from lower numbers to 8 leads to consistent performance improvements. Using bag size 16 has further slight performance boost. We report our main results with a bag size of 16. 

% \begin{table*}[!tb]%[!htbp]
% \small
% \centering
% % \begin{tabular}{@{}c M{0.8cm} M{0.8cm} M{0.8cm} M{0.7cm} M{0.7cm}M{0.7cm}@{}}
% % \begin{tabular}{@{}l @{\hspace{2em}}cccccccccccccc}
% \begin{tabular}{ccccccccccccccc}
% \toprule
% % Model & \multicolumn{3}{c}{TANet} & \multicolumn{3}{c}{Swin} \\
% % \cmidrule(r){2-4}\cmidrule{5-7}
% % Dataset & \multirow{2}*{pretrain} & \multirow{2}*{sett.} & \multicolumn{4}{c}{HMDB51}  & \multicolumn{4}{c}{UCF101} & \multicolumn{4}{c}{SSv2}  \\
% % \cmidrule(r){4-7}\cmidrule(r){8-11}\cmidrule(r){12-15}
% Bag size & UCF101 & HMDB51 & K600 & MiniSSv2 & Charades & UAV Human & Moments-in-time \\
% \midrule
% 1 & 77.26 & 49.85 & 70.08 \\
% 4 & 77.24 & 49.84 & 70.71 & 5.09 & 23.49 & 2.46 & 21.84\\
% 8 & 77.70 & 51.61 & 71.35 & 6.50 & 23.38 & 2.49 & 23.03\\
% 16 & 77.88 & 51.09 & 71.24 & 5.46 & 23.52 & 2.53 & 22.44\\

% \bottomrule
% \end{tabular}
% \caption{
% Bag size. We use text bags of GPT3 verbs + BLIP verbs. 100\% pseudo labels.
%  }
% \label{tab:bag_size}
% % \label{tab:yti_state_of_the_art}
% \end{table*}

\begin{table}[!tb]%[!htbp]
\footnotesize
\centering
% \begin{tabular}{@{}c M{0.8cm} M{0.8cm} M{0.8cm} M{0.7cm} M{0.7cm}M{0.7cm}@{}}
% \begin{tabular}{@{}l @{\hspace{2em}}cccccccccccccc}
\begin{tabular}{cccc}
\toprule
% Model & \multicolumn{3}{c}{TANet} & \multicolumn{3}{c}{Swin} \\
% \cmidrule(r){2-4}\cmidrule{5-7}
% Dataset & \multirow{2}*{pretrain} & \multirow{2}*{sett.} & \multicolumn{4}{c}{HMDB51}  & \multicolumn{4}{c}{UCF101} & \multicolumn{4}{c}{SSv2}  \\
% \cmidrule(r){4-7}\cmidrule(r){8-11}\cmidrule(r){12-15}
Bag size & UCF101 & HMDB51 & K600  \\
\midrule
1 & 77.26 & 49.85 & 70.08 \\
4 & 77.24 & 49.84 & 70.71 \\
8 & \underline{77.70} & \underline{50.61} & \textbf{71.35} \\
16 & \textbf{77.88} & \textbf{51.09} & \underline{71.24} \\

\bottomrule
\end{tabular}
% \vspace{-2mm}
\caption{
Effect of bag size. We report the zero-shot transfer performance on UCF, HMDB and K600.
For a thorough ablation, we set the text bag filtering ratio $p=100\%$ to keep the full noisy text bag property.
 }
\label{tab:bag_size}
% \label{tab:yti_state_of_the_art}
\end{table}
\section{Conclusion}
In this work, we consider the task of leveraging unlabeled video collections and a set of language sources to finetune the \vl{} model for improved zero-shot action recognition. To our best knowledge, our approach `\oursfull{}' (\OurMethod) is the first of this kind. Specifically, we leverage a set of language sources (unpaired action dictionaries, Large Language Models and VL models) to construct a text bag for each unlabeled video. Then we use the unlabeled videos and text bags to finetune the \vl{} model with the objective of Multiple Instance Learning. Our extensive evaluation for zero-shot and few-shot action recognition across several unseen action benchmarks demonstrate significant performance improvement over the source \vl{} model, as well as improvement over baselines trained in a fully supervised manner.

% \newpage
\appendix
\section*{Supplementary}
For further insights into our approach \OurMethod, we introduce more dataset statistics (Sec.~\ref{sec:dataset_statistics}) and implementation details (Sec.~\ref{sec:implementation_detials}) of \OurMethod. 

In the additional results, we provide comparison of visualizations of attention heatmaps across several approaches in Sec.~\ref{sec:attention-heatmaps}. Furthermore, we report more results of finetuning with noisy action dictionary (Sec.~\ref{sec:robustness_noisy_dictionary_more}), and provide more examples of language sources used for training (Sec.~\ref{sec:examples_of_language_sources}). Lastly, we explore a cross-frame attention temporal module in Sec.~\ref{sec:parameter_free_temp_module}. 

% \section{Source Code}\label{sec:source_code}
% We provide the source code in the directory \verb$supplementary_3246/source_code/MAXI$. Check \verb$README.md$ for instructions of re-evaluation. The source code will be released upon acceptance. 

\begin{figure*}[ht]
\includegraphics[width=\columnwidth]{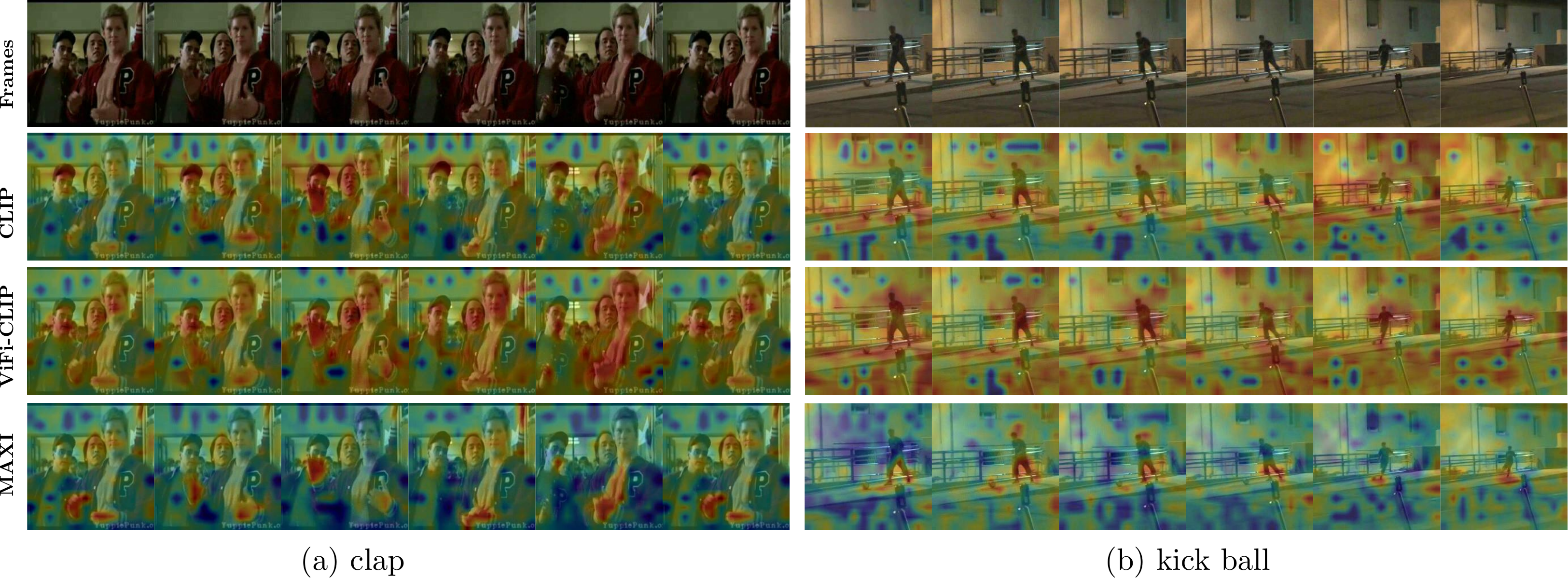}
\caption{
\label{fig:supp_attn_in_dict_samples}
Attention heatmaps on actions which have a verb form (lemma or gerund) directly included in the K400 dictionary. We compare among CLIP (2nd row), ViFi-CLIP (3rd row) and our \OurMethod (4th row). Warm and cold colors indicate high and low attention. \OurMethod has more focused attention on hands (for \textit{clap}) and legs (for \textit{kick ball}). 
}
\end{figure*}

\section{Dataset Statistics}\label{sec:dataset_statistics}
\textbf{Kinetics-400} (K400)~\cite{kay2017kinetics}  is the most popular benchmark for action recognition tasks, containing around 240K training videos in 400 classes. The dataset consists of YouTube videos with an average length of 10 seconds. We use the training set of K400 for finetuning CLIP. 

\textbf{UCF101}~\cite{soomro2012ucf101} is collected from YouTube videos, consisting of 13K videos from 101 classes. There are three splits of training data ($\sim$ 9.4K) and validation data ($\sim$3.6K). Following XLCIP~\cite{ni2022expanding} and ViFi-CLIP~\cite{rasheed2022fine}, we report the average performance on the three validation splits. 

\textbf{HMDB51}~\cite{kuehne2011hmdb} consists of 7K videos comprised of 51 action classes, collected from YouTube videos and movie clips. There are three splits of training data ($\sim$ 3.5K, 70 videos per class) and validation data ($\sim$1.5K, 30 videos per class). Following~\cite{ni2022expanding,rasheed2022fine}, we report the average performance on the three validation splits. 

\textbf{Kinetics-600} (K600)~\cite{carreira2018short} is an extension of K400, consisting of 650K videos in 600 classes. Following~\cite{chen2021elaborative, ni2022expanding, rasheed2022fine}, we use the 220 classes\footnote{In the evolution from K400 to K600, there are renamed, removed and split classes. See details in Appendix B in~\cite{chen2021elaborative}.} that are not included in K400 for zero-shot action recognition. There are three validation splits, each containing 160 classes randomly sampled from these 220 classes. We report the average performance on the three validation splits, each containing around 14K videos. 

\textbf{MiniSSv2}~\cite{chen2021deep} (87 classes, 93K videos) is a subset of Something-Something v2 (SSv2)~\cite{goyal2017something} (174 classes, 220K videos). SSv2 is an egocentric motion-based action dataset, which has a large visual domain shift to K400. Furthermore, the action classes are detailed descriptions of fine-grained movements, in a largely different language style than the K400 action dictionary, \eg \textit{Failing to put something into something because something does not fit}, and \textit{Lifting a surface with something on it but not enough for it to slide down}. For zero-shot action recognition, we evaluate on the validation split of MiniSSv2 (12K videos). For few-shot action recognition, we follow~\cite{rasheed2022fine} and evaluate on the validation split of SSv2 (25K videos).

\textbf{Charades}~\cite{sigurdsson2016hollywood} is a long-range activity dataset recorded by people in their homes based on provided scripts for home activities. There are $\sim$10K videos in 157 classes. The average video length is 30 seconds. Each video has annotations of an average of 6.8 action instances, often in complex co-occurring cases. The validation split consists of 1.8K videos. We report the mean Average Precision (mAP) for the multi-label classification task. 

\textbf{Moments-in-Time} (MiT)~\cite{monfort2019moments} is a large-scale action dataset of 3-second YouTube video clips, which cover actions in 305 classes, performed by both humans and animals. The validation split consists of 30K videos. 

\textbf{UAV Human} (UAV)~\cite{li2021uav} is an action dataset recorded with an Unmanned Aerial Vehicle in unique camera viewpoints. There are 155 action classes. Actions in different categories are performed by a fixed group of subjects in the same background scenes. This leads to an extremely low object-scene bias and a large shift to the domain of K400 and CLIP. We evaluate on the RGB videos and report the average performance on the two official validation splits, each consisting of $\sim$ 6.2K videos. 

% \wei{explain the issue with UCF class name, with space} we add space to have human-readable class names on UCF101.

\begin{figure*}[ht]
\includegraphics[width=\columnwidth]{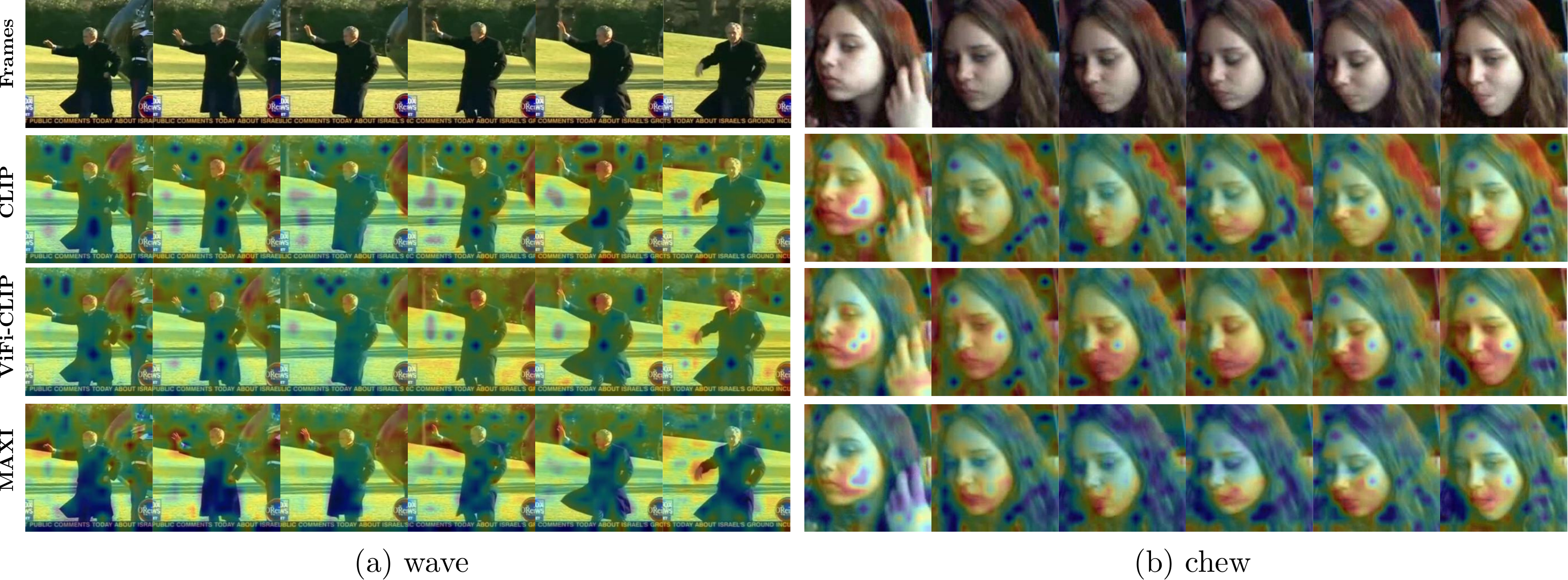}
\caption{
\label{fig:supp_attn_out_of_dict_samples}
Attention heatmaps on novel actions which do not have any verb form included in the K400 dictionary. We compare among CLIP (2nd row), ViFi-CLIP (3rd row) and our \OurMethod (4th row). Warm and cold colors indicate high and low attention. \OurMethod has more focused attention on hand and arm for \textit{wave}, and on the area of mouth for \textit{chew}.
}
\end{figure*}

\begin{figure*}[ht]
\includegraphics[width=\columnwidth]{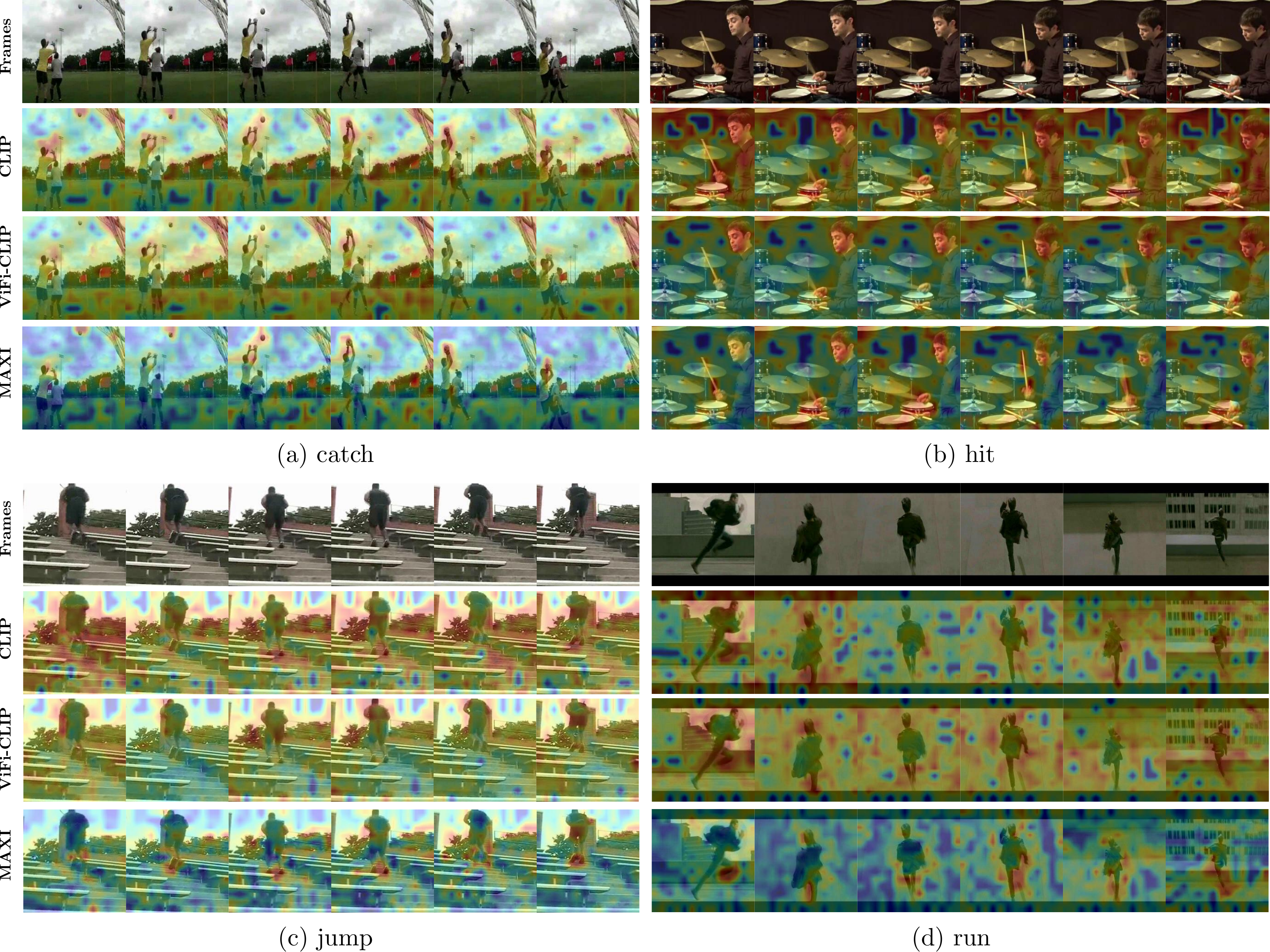}
\caption{
\label{fig:supp_attn_general_action_samples}
Attention heatmaps on actions which have a verb form (lemma or gerund) directly included in the K400 dictionary. We compare among CLIP (2nd row), ViFi-CLIP (3rd row) and our \OurMethod (4th row). Warm and cold colors indicate high and low attention. \OurMethod has more concentrated attention on the part where the action happens, \eg catching ball with hands (Fig.~\ref{fig:supp_attn_general_action_samples}(a), 4th row), hitting drum with stick (Fig.~\ref{fig:supp_attn_general_action_samples}(b), 4th row), legs and feet jump on stairs (Fig.~\ref{fig:supp_attn_general_action_samples}(c), 4th row), and attention on the running body (Fig.~\ref{fig:supp_attn_general_action_samples}(d), 4th row).
}
\end{figure*}

\section{Implementation Details}\label{sec:implementation_detials}
In addition to the details mentioned in the main manuscript, we cover more implementation specifics here. 

\textbf{CLIP matching.} 
The CLIP matching step is for consuming the language source of the predefined action dictionary $D$. 
We use CLIP\footnote{CLIP model \href{https://openaipublic.azureedge.net/clip/models/5806e77cd80f8b59890b7e101eabd078d9fb84e6937f9e85e4ecb61988df416f/ViT-B-16.pt}{source} }~\cite{clip} with the ViT-B/16 visual encoder~\cite{dosovitskiy2021image} to match each video with texts in the predefined action dictionary. To improve the matching quality for Text Bag Construction, we perform prompt ensembling over the 28 prompt templates\footnote{\url{https://github.com/openai/CLIP/blob/main/data/prompts.md}} which are proposed by CLIP for Kinetics videos. Important to note, during inference we follow the exact protocol of ViFi-CLIP \cite{rasheed2022fine} and use only a single prompt.
% . 

\textbf{GPT-3 text expansion.}
We employ the GPT-3 \texttt{text-davinci-003} model~\cite{brown2020language}. We set the temperature to 0.4. We generate 5 verb phrases using the input instruction - \textit{Generate 5 phrases to describe the action of $<$action$>$ in simple words}. Here for a video $x_i$, \textit{$<$action$>$} is the best matched text $\hat t_i$ from the predefined action dictionary. 

\textbf{BLIP captioning.}
We use BLIP model \footnote{BLIP model \href{https://storage.googleapis.com/sfr-vision-language-research/BLIP/models/model_large_caption.pth}{source} }~\cite{blip} with ViT-L/16 as the image captioner. For each video, the image captioning is performed on 8 uniformly sampled frames. The frames are resized into 384.

\textbf{Text augmentation}. We use the natural language processing tool spaCy\footnote{spaCy \url{https://spacy.io/}} to parse the verbs and verb phrases from the descriptions. We perform augmentation by converting the verbs into forms of lemma and gerund (present participle) and include results in the text bag. 

\textbf{Training.} We employ CLIP with the ViT-B/16 visual encoder. We follow the full-finetuning configuration of~\cite{rasheed2022fine} to finetune both the visual and text encoder. 
During training, we follow the configuration of~\cite{rasheed2022fine,ni2022expanding} for visual augmentation of multi scale crop, random flipping, color jitering and gray scaling. We do not perform augmentations of MixUp or CutMix. 

As different videos have varying numbers of texts in their bags, we randomly sample $N_{\text{bag}}$ texts from the originally constructed bag in each training iteration. 
For multiple instance learning, we use all the $N_{\text{bag}}$ words in a text bag to form $N_{\text{bag}}$ text prompts for each video. 
The text prompt is in the format of \textit{$<$text1$>$ + $<$text2$>$}.
The first part \textit{$<$text1$>$} is uniform for all the $N_{\text{bag}}$ text prompts. Specifically, we use a hand-crafted prompt template \textit{a photo of $<$action$>$}, where \textit{$<$action$>$} is the best-matched text $\hat t_i$ from the predefined action dictionary (see Eq.~1 in the main manuscript). \textit{$<$text2$>$} is an individual text from the text bag. 
To avoid duplication, we do not use $\hat t_i$ as \textit{$<$text2$>$}.
% For \textit{$<$text2$>$}, we do not consider $\hat t_i$ in the text bag to avoid duplication. 

\textbf{Inference.} We follow~\cite{ni2022expanding,rasheed2022fine} and sample a single view via sparse temporal sampling and spatial center crop. The same single prompt template is used in inference. 

% Train on 4$\times$ A6000

\begin{table*}[!tb]%[!htbp]
\scriptsize
\centering
% \begin{tabular}{@{}c M{0.8cm} M{0.8cm} M{0.8cm} M{0.7cm} M{0.7cm}M{0.7cm}@{}}
% \begin{tabular}{@{}l @{\hspace{2em}}cccccccccccccc}
\begin{tabular}{ccccccccccccccc}
\toprule
Action dictionary & dictionary size & UCF101 & HMDB51 & K600 & MiniSSv2 & Charades & UAV Human & Moments-in-time \\
\midrule
\multicolumn{2}{c}{CLIP~\cite{clip} (w/o finetune) Zero-Shot} & 69.93 / 92.7 & 38.02 / 66.34 & 63.48 / 86.80 & 3.96 / 14.42 & 19.80 & 1.79 / 7.05 & 20.11 / 40.81 \\
\midrule
K400 & 400 & \textbf{78.18} / \textbf{96.03} & \textbf{50.35} / \textbf{77.10} &  \textbf{70.78} / \textbf{92.17} & \textbf{5.74} / \textbf{17.70} & \textbf{23.89} & \textbf{3.06} / \textbf{9.46} & \textbf{22.41} / \textbf{45.83} \\
% MiniKinetics & 200 & 75.10 / 95.82 & \underline{48.34} / \underline{76.95} & \underline{69.23} / 90.92 & \textbf{6.50} / \textbf{18.76} & \underline{22.70} & \underline{2.40} / \underline{8.04} & \textbf{22.50} / \textbf{46.01} \\
K400+WebVid2.5M & 800 & 75.99 / 96.00 & \underline{45.97} / 73.94 & 69.14 / \underline{91.13} & \underline{4.81} / \underline{15.79} & 22.67 & 2.11 / \underline{8.00} & 20.92 / 43.99 \\
K400+WebVid2.5M & 1200 & 75.72 / \underline{96.02}  & 45.51 / \underline{73.97} & \underline{69.36} / 91.11 & 4.21 / 15.15 & 22.35 & \underline{2.39} / 7.98 & \underline{21.29} / \underline{44.33}  \\
K400+WebVid2.5M & 1600 & \underline{76.14} / 96.01 & 44.84 / 71.79 & 69.23 / 91.10 & 4.42 / 14.71 & \underline{22.89} & 2.14 / 7.71 & 20.69 / 43.59 \\
% UCF101 & 101 & -  \\

\bottomrule
\end{tabular}
\vspace{-2mm}
\caption{
Robustness of finetuning with noisy action dictionaries. We add noisy verbs parsed from the WebVid2.5M dataset into the original K400 action dictionary. We report the zero-shot transfer performance (mAP on Charades and Top1/Top5 accuracy on other datasets). We set the text bag filtering ratio $p=50\%$ for improved text bag quality.
}
\label{tab:supp_noisy_class_space}
% \label{tab:yti_state_of_the_art}
\end{table*}

\begin{figure*}[ht]
\includegraphics[width=\columnwidth]{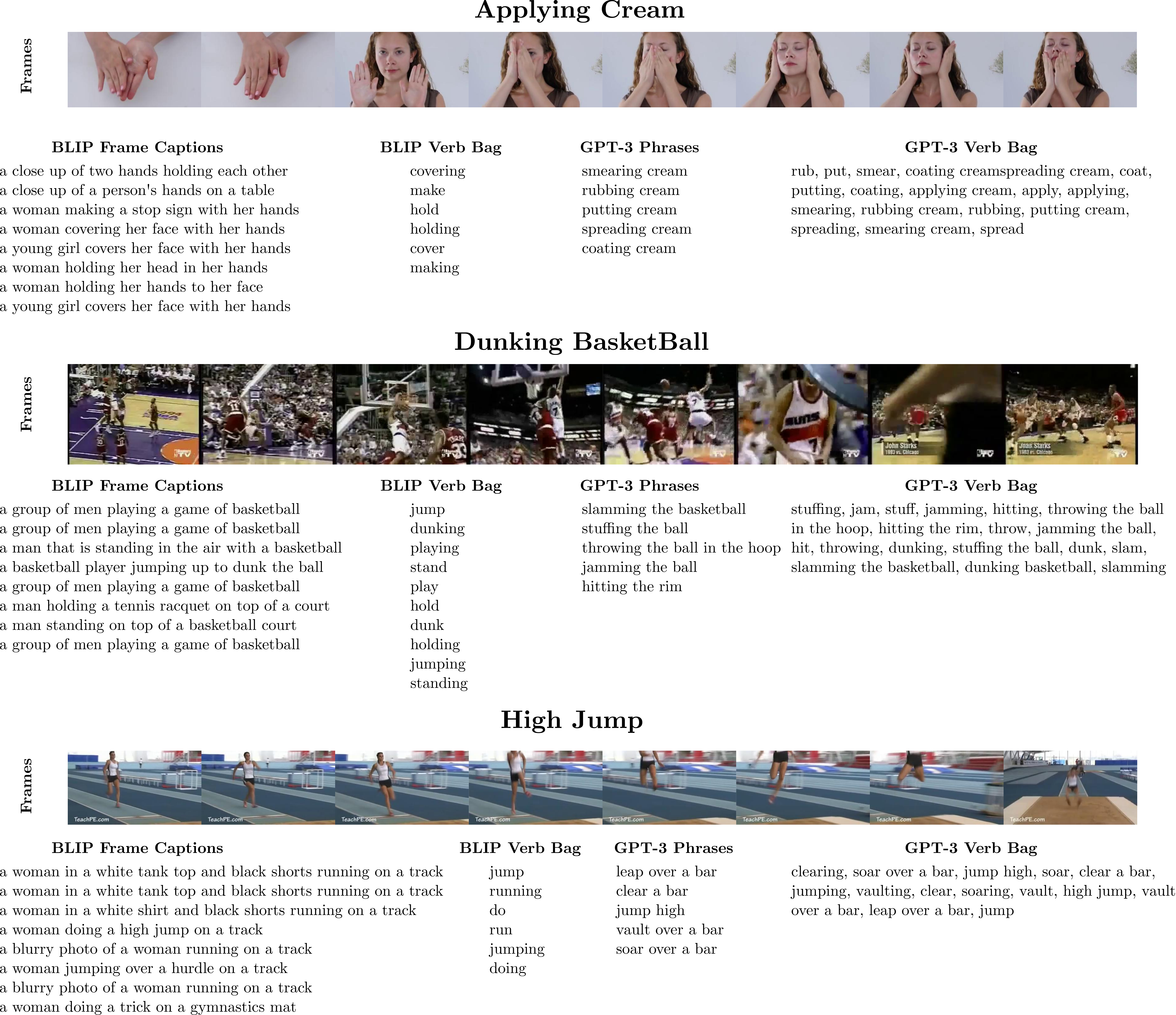}
\caption{
\label{fig:caption_gpt_examples}
Examples of video frames, BLIP frame captions, GPT-3 phrases, together with the derived BLIP verb bag and GPT-3 verb bag. The videos are from the K400 dataset. 
}
\end{figure*}

\begin{table*}[!tb]%[!htbp]
\footnotesize
\centering
% \begin{tabular}{@{}c M{0.8cm} M{0.8cm} M{0.8cm} M{0.7cm} M{0.7cm}M{0.7cm}@{}}
% \begin{tabular}{@{}l @{\hspace{2em}}cccccccccccccc}
\begin{tabular}{ccccccccccccccc}
\toprule
Temp. attention layers & UCF101 & HMDB51 & K600 & MiniSSv2 & Charades & UAV Human & Moments-in-time \\
% \midrule
% \multicolumn{2}{c}{CLIP~\cite{clip} (w/o finetune) Zero-Shot} & 69.93 / 92.7 & 38.02 / 66.34 & 63.48 / 86.80 & 3.96 / 14.42 & 19.80 & 1.79 / 7.05 & 20.11 / 40.81 \\
\midrule
None & \textbf{78.17} & \textbf{52.24} & \textbf{71.43} & \textbf{6.37} & \textbf{23.79} & \underline{2.72} & \textbf{22.91}  \\
% MiniKinetics & 200 & 75.10 / 95.82 & \underline{48.34} / \underline{76.95} & \underline{69.23} / 90.92 & \textbf{6.50} / \textbf{18.76} & \underline{22.70} & \underline{2.40} / \underline{8.04} & \textbf{22.50} / \textbf{46.01} \\
2 & \underline{77.38} & 51.83 & \underline{70.41} & 5.98 & \underline{22.87} & \textbf{2.90} & 22.51 \\
6 & 75.91 & \underline{51.92} & 69.23 & \underline{6.09} & 21.78 & 2.52 & \underline{22.52}  \\
% UCF101 & 101 & -  \\

\bottomrule
\end{tabular}
\vspace{-2mm}
\caption{
Cross-frame temporal attention modules. we report the zero-shot transfer performance after finetuning CLIP on K400. We train with text bags of GPT3 verbs and BLIP verbs. We set the text bag filtering ratio $p=90\%$. Adding temporal attention module does not lead to performance improvement. 
}
\label{tab:supp_tempt_att}
% \label{tab:yti_state_of_the_art}
\end{table*}

\section{Additional Results}

\subsection{Attention Heatmaps}\label{sec:attention-heatmaps}
To gain more insights into the performance improvement of \OurMethod, we compare the visualizations of attention heatmaps across several approaches in Fig.~\ref{fig:supp_attn_in_dict_samples}, Fig.~\ref{fig:supp_attn_out_of_dict_samples} and Fig.~\ref{fig:supp_attn_general_action_samples}. \textit{CLIP} is the original CLIP~\cite{clip} without any finetuning. \textit{ViFi-CLIP}~\cite{rasheed2022fine} finetunes CLIP via supervised classification on K400 with ground truth annotations. \textit{\OurMethod} is our approach of unsupervised finetuning with language knowledge. 

We obtain the attention maps by computing the cosine similarity between the patch token features from the visual encoder and the text feature from the text encoder. We visualize the attention maps in several action classes from the downstream datasets used for the zero-shot action recognition task. Based on the relationship between the zero-shot action class and the K400 action dictionary used for training, we categorize the visualizations into 3 groups: (1) In-dictionary action classes which have a verb form (lemma or gerund) directly included in the K400 action dictionary, \eg \textit{clap} and \textit{kick ball} in Fig.~\ref{fig:supp_attn_in_dict_samples}; (2) Novel actions classes which do not have any verb form included in the K400 action dictionary, \eg \textit{wave} and \textit{chew} in Fig.~\ref{fig:supp_attn_out_of_dict_samples}; (3) General actions whose verb form is a basic component of several actions in the K400 action dictionary, \eg \textit{catch}, \textit{hit}, \textit{jump} and \textit{run} in Fig.~\ref{fig:supp_attn_general_action_samples}.

\myparagraph{In-dictionary action classes.} In Fig.~\ref{fig:supp_attn_in_dict_samples}, we visualize two samples of the action \textit{clap} and \textit{kick ball}. \textit{clap} has the same lemma as \textit{clapping} in the K400 dictionary, while \textit{kick ball} has related actions of \textit{kicking field goal} and \textit{kicking soccer ball} in the K400 dictionary. We see that CLIP has incorrectly high attention on object (Fig.~\ref{fig:supp_attn_in_dict_samples}(a), 2nd row) or background scene (Fig.~\ref{fig:supp_attn_in_dict_samples}(b), 2nd row). ViFi-CLIP has cluttered high attention on both the subjects and the background scenes. On the contrary, \OurMethod has more focused attention on the hands (for \textit{clap}) and legs (for \textit{kick ball}). 

In our GPT-3 text bag of \textit{clapping}, related words such as \textit{clap}, \textit{smacking hands}, \textit{slapping palms} and \textit{clapping hands} are included. This strengthens the association between the action \textit{clap} and the body part of hands, and leads to more accurate attention. Furthermore, in BLIP caption verb text bags, the verb \textit{clap} appears several times in frame captions of K400 videos of \textit{clapping}, \textit{giving or receiving award} and \textit{applauding}. This further improves the understanding of \textit{clap}. Similarly, in BLIP frame captions, \textit{kick} is an even more basic verb with large amount of occurrences.

\myparagraph{Novel action classes.} In Fig.~\ref{fig:supp_attn_out_of_dict_samples}, we compare the attention maps for the novel verbs \textit{wave} and \textit{chew} that do not appear in the K400 action dictionary. We see that for \textit{wave}, CLIP and ViFi-CLIP have attention on the background scene or on the head, while \OurMethod has correct attention on the hand and arm. For \textit{chew}, CLIP has more attention on the hair and ViFi-CLIP has attention on a large area of the face. On the contrary, \OurMethod has consistent focused attentions on the area of the mouth where the action  \textit{chew} happens. 

The verb \textit{wave} appears in BLIP caption verb text bags of several K400 videos of \textit{clapping}, \textit{applauding}, \textit{celebrating}. The verb \textit{chew} appears in captions of K400 videos of \textit{eating carrots}, \textit{eating spaghetti}, \textit{eating watermelon} and \textit{baby waking up}. The additional language source improves the knowledge of actions that never appear in the K400 action dictionary. 

\myparagraph{General actions.} In Fig.~\ref{fig:supp_attn_general_action_samples}, we illustrate the attention maps for four general verbs \textit{catch}, \textit{hit}, \textit{jump} and \textit{run}. These verbs are basic components of several actions in the K400 dictionary, \eg \textit{catching fish}, \textit{catching or throwing frisbee}, \textit{hitting baseball}, \textit{jumping into pool} and \textit{running on treadmill}. In these samples, CLIP and ViFi-CLIP have cluttered attention on the background scene or objects. \OurMethod has more concentrated attention on the part where the action happens, \eg catching ball with hands (Fig.~\ref{fig:supp_attn_general_action_samples}(a), last row), hitting drum with stick (Fig.~\ref{fig:supp_attn_general_action_samples}(b), last row), legs and feet jump on stairs (Fig.~\ref{fig:supp_attn_general_action_samples}(c), last row), and attention on the running body (Fig.~\ref{fig:supp_attn_general_action_samples}(d), last row).

These verbs are very general and could have highly diverse instantiations. \Eg \textit{hit} (drum) in Fig.~\ref{fig:supp_attn_general_action_samples}(b) is not close to \textit{hitting baseball} on K400. \textit{jump} (on stairs) in Fig.~\ref{fig:supp_attn_general_action_samples}(c) is not close to \textit{jumping into pool} or \textit{bungee jumping} on K400, even if they share the same verb. In our GPT-3 verb bag and BLIP caption verb bag, there is a large amount of these verb instances that facilitate the comprehensive understanding of these general verbs. This leads to better focus even in unusual complex scenes, \eg jumping on stairs (Fig.~\ref{fig:supp_attn_general_action_samples}(c)).

% CLIP has object bias and tends to have attention on the distinct objects. 

\subsection{Robustness Against Noisy Action Dictionary}\label{sec:robustness_noisy_dictionary_more}
In Table~5 in the main manuscript, we explored the robustness of our finetuning pipeline against noisy action dictionaries. In case of an over-specified dictionary, we added noisy verbs and verb phrases into the original K400 action dictionary. The noisy verbs are parsed from the captions in the WebVid2.5M dataset~\cite{bain2021frozen}. Here we further increase the ratio of noisy verbs, and add 800 and 1200 verbs into the dictionary, resulting in 1200-class and 1600-class spaces. 

In Table~\ref{tab:supp_noisy_class_space}, we report the zero-shot transfer performance of models finetuned with the resulted 1200-class and 1600-class space. We set the text bag filtering ratio $p=50\%$ for improved text bag quality. We see that even with extremely noisy dictionary where 50\% to 75\% of words do not match with the video data, our finetuning still results in a robust zero-shot transfer performance to unseen datasets. 
The robustness is the consequence of the fact that we collect knowledge from multiple language sources and learn from them via Multiple Instance Learning.
Note that the zero-shot transfer does not have consistent change in performance across the downstream datasets, as different datasets have different language domain shift to the action dictionary used for training.

\subsection{Examples of Language Sources}\label{sec:examples_of_language_sources}
Similar to the \textit{cooking egg} example in Fig.~2 in the main manuscript, we illustrate more examples of video frames, BLIP frame captions, GPT-3 phrases, together with the derived BLIP verb bag and GPT-3 verb bag in Fig.~\ref{fig:caption_gpt_examples}. The videos are from the unlabeled K400 dataset which we use for training.

\subsection{Parameter-Free Temporal Module}\label{sec:parameter_free_temp_module}
As mentioned in Sec.~3.1 in the main manuscript, we explore a parameter-free temporal-aware module on the CLIP model. We modify the multi head attention module~\cite{vaswani2017attention} in the visual encoder of CLIP to be temporal aware. Originally, the attention on the frame $t$ is computed via 
$A_t(Q_t, K_t, V_t)= \text{softmax}\frac{Q_t {K_t}^\top }{d_k}V_t$, where $Q_t$, $K_t$ and $V_t$ are the query, key and value from frame $t$. 

We explore to compute the cross-frame attention via 
\begin{equation}
\small
A'_t(Q_t, K_{t+i}|_{i\in I}, V_t)= 
\text{softmax}
\frac{ \sum_{i\in I} (Q_t \cdot {K_{t+i}}^\top) / |I|}
{d_k}V_t
\end{equation}
where we set $I=\{-1, 0, 1\}$. In this case, we use the keys from the frame $t-1$, $t$ and $t+1$ to compute the attention for frame $t$. 

We apply the cross-frame attention on the last 2 and on the last 6 transformer layers in the visual encoder of CLIP. In Table~\ref{tab:supp_tempt_att}, we report the zero-shot transfer performance. We see that in comparison to the variant without any temporal attention module, using cross-frame attention does not lead to performance improvement. K400 is of far smaller scale in comparison to the original CLIP domain. Finetuning from the CLIP model weights with a modified architecture could result in the case that the model drifts far away from the wise CLIP source domain. The results are consistent with the claims in \cite{rasheed2022fine} that a sophisticated temporal module does not necessarily lead to performance improvement.
% (2) K400 contains YouTube videos where the temporal reasoning does not have a big impact on recognizing actions. 

% \textbf{We do not add additional parameters onto CLIP}. 

% Explanation: K400 is of a lot smaller scale in comparison to the original CLIP source domain. as we finetune the entire model, optimization is sensitive, the model could easily drift away. on online user videos, the temporal reasoning does not have a big impact. 

{\small
\bibliographystyle{ieee_fullname}
\bibliography{egbib}
}

\end{document}